\documentclass[pdflatex,sn-mathphys-num,iicol]{sn-jnl}
\usepackage{graphicx} 
\usepackage{makecell}
\usepackage{multirow}
\usepackage[table]{xcolor}
\usepackage[title]{appendix}%
\usepackage{float}  
\usepackage{amsmath}
\usepackage{bm}
\usepackage{paralist}
\usepackage{microtype}

\DeclareMathOperator{\odesolv}{ODESolver}
\DeclareMathOperator{\mlp}{MLP}
\DeclareMathOperator{\gru}{GRU}
\DeclareMathOperator{\mhsa}{MHSA}
\DeclareMathOperator{\mhca}{MHCA}
\DeclareMathOperator{\diag}{diag}

\renewcommand{\orcidlogo}{%
  \includegraphics[width=7pt]{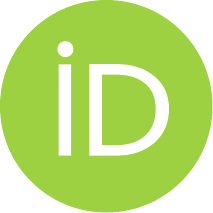}%
}

\begin{document}

\title[Article Title]{Beyond Kalman Filters: Deep Learning-Based Filters for Improved Object Tracking}

\author[1]{\fnm{Momir} \sur{Adžemović}}\email{mi17042@alas.matf.bg.ac.rs}

\author[2]{\fnm{Predrag} \sur{Tadić} \orcid{0000-0002-8845-997X}}\email{ptadic@etf.bg.ac.rs}

\author[3]{\fnm{Andrija} \sur{Petrović}}\email{andrija.petrovic@fon.bg.ac.rs}

\author[1]{\fnm{Mladen} \sur{Nikolić} \orcid{0009-0002-8943-2709}}\email{mladen.nikolic@matf.bg.ac.rs}

\affil[1]{\orgdiv{Department of Computer Science}, \orgname{Faculty of Mathematics, University of Belgrade}, \orgaddress{\city{Belgrade}, \country{Serbia}}}

\affil[2]{\orgdiv{Department of Signals and Systems}, \orgname{School of Electrical Engineering, University of Belgrade}, \orgaddress{\city{Belgrade}, \country{Serbia}}}

\affil[3]{\orgdiv{Center for Bussiness Decision Making}, \orgname{Faculty of Organizational sciences, University of Belgrade}, \orgaddress{\city{Belgrade}, \country{Serbia}}}

\date{February 2024}

\abstract{Traditional tracking-by-detection systems typically employ Kalman filters (KF) for state estimation. However, the KF requires domain-specific design choices and it is ill-suited to handling non-linear motion patterns. To address these limitations, we propose two innovative data-driven filtering methods. Our first method employs a Bayesian filter with a trainable motion model to predict an object's future location and combines its predictions with observations gained from an object detector to enhance bounding box prediction accuracy. Moreover, it dispenses with most domain-specific design choices characteristic of the KF. The second method, an end-to-end trainable filter, goes a step further by learning to correct detector errors, further minimizing the need for domain expertise. Additionally, we introduce a range of motion model architectures based on Recurrent Neural Networks, Neural Ordinary Differential Equations, and Conditional Neural Processes, that are combined with the proposed filtering methods. Our extensive evaluation across multiple datasets demonstrates that our proposed filters outperform the traditional KF in object tracking, especially in the case of non-linear motion patterns---the use case our filters are best suited to. We also conduct noise robustness analysis of our filters with convincing positive results. We further propose a new cost function for associating observations with tracks. Our tracker, which incorporates this new association cost with our proposed filters, outperforms the conventional SORT method and other motion-based trackers in multi-object tracking according to multiple metrics on motion-rich DanceTrack and SportsMOT datasets.}

\keywords{Filter, Motion model, Deep Learning, Multi-object tracking, Neural Ordinary Differential Equations, Conditional Neural Processes}

\maketitle

\section{Introduction}

Object tracking is a critical component in various applications, such as autonomous vehicle driving~\cite{object_tracking_in_autunomous_driving, nuScenes}, robotics~\cite{object_tracking_in_robotics}, and surveillance~\cite{object_tracking_in_surveillence}, where it's essential for understanding and predicting the movements and behaviors of objects in real-time. 
The tracking-by-detection paradigm, which involves associating detected object bounding boxes over successive frames, has been propelled by the rise of fast and accurate deep learning-based object detection models~\cite{yolox, yolov8, faster_rcnn, detr}. This approach initially detects objects in each video frame and then associates these detected objects across adjacent frames. An essential part of most association methods is the motion-based association, which relies on the bounding box geometric features and motion models predicting the object’s movement patterns~\cite{sort}. 

Classical algorithms for motion-based association rely on the Kalman Filter (KF)~\cite{sort, deepsort, jde, botsort}. The KF is a strong baseline in the case of predominantly linear motion~\cite{sort, mot20}, but fails to handle non-linear motion effectively~\cite{dancetrack}. Some alternatives~\cite{pf,ekf,ukf} perform better for non-linear motion modeling, but they cannot directly learn motion patterns from data and do not figure prominently in object tracking. Both the KF and these alternatives require domain knowledge for heuristics design and careful tuning of the hyperparameters. 

To address these drawbacks of the existing approaches, we propose two new filtering methods that leverage non-linear, learnable probabilistic motion models. These filters are designed to directly replace the KF in any tracking solution that employs it and they offer a lot of flexibility in the selection of motion model architectures. In contrast to the KF, our filters rely on deep-learning motion models that are drastically less dependent on domain knowledge, since they are learned from the data. The first filtering method uses a learned motion model to estimate a distribution over the object's future position. Then, it uses Bayesian inference to combine this prediction with a new detection. Unlike the KF which requires domain-specific heuristics and hyperparameters for initialization, motion model, process noise, and measurement noise, our Bayesian filters only necessitate a heuristic for approximating the measurement (detector) noise. In the second filtering method, a model learns to predict the object's future position and to combine it with the detector measurement in an \textit{end-to-end} fashion. Thus it eliminates the measurement noise hyperparameters which are needed in the first approach. Due to the proposed training methodology, once the filter is trained, the object detector can be replaced without any need for further adjustments or hyperparameter tuning, providing more flexibility in deployment.

As a core component of these filters, we propose deep learning probabilistic models capable of directly learning non-linear motion patterns from data. Moreover, these models can learn dataset-specific characteristics, such as aspect ratios of objects in pedestrian datasets or frequent object positioning and motion patterns. We opt for Neural ODEs~\cite{node} and Neural Processes~\cite{np, cnp} due to their robustness against missing and noisy data.

Further, we improve the tracker logic by proposing a modification to the commonly used Intersection Over Union (IoU) association cost~\cite{sort, bytetrack}. IoU is a good baseline, but it does not consider the object's scale and position which is important during occlusions. Therefore, we combine IoU with the L1 distance between the scale and position of predicted versus detected bounding boxes, lowering the occurrence of ID switches.

In this work, we simplify our tracking algorithm by focusing solely on motion-based association, similar to the SORT method, and avoid using visual features~\cite{deepsort}. This approach enables us to better isolate the contributions of the proposed techniques to the final tracker performance and to evaluate them more precisely.

We summarize our contributions as follows:
\begin{itemize}
    \item We propose a non-linear Bayesian filter, which can incorporate non-linear motion models into the Bayesian filtering approach. This filter is a direct alternative to the standard KF in the tracking-by-detection paradigm.
    \item We propose an end-to-end learnable filter for object tracking, capable of adapting to an arbitrary detector without the need for additional training or hyperparameter adjustments.
    \item We dispense with the domain-specific design choices of the KF, by allowing them to be learned from data.
    \item We consider several deep non-linear motion models designed for trajectory extrapolation with uncertainty estimates that fit the filtering paradigm.
    \item We propose a simple, motion-based association cost function designed for tracking multiple objects undergoing extensive movement and occlusions. 
    \item We build MoveSORT, a SORT-like tracker using our filtering and association methods. 
    \item We conduct a detailed analysis of the effectiveness of the proposed filters and derived trackers across multiple object-tracking datasets. They significantly outperform the KF on dynamic datasets characterized by non-linear motion. Their performance on datasets where linear motion dominates is comparable to that of the KF.
\end{itemize}

\section{Background}
\label{sec:background}

In this section, we provide a concise overview of the prerequisites necessary to understand the proposed methodology. Specifically, we briefly cover the tracking-by-detection paradigm and the deep learning architectures on which we base our motion models.

\subsection{Tracking-by-detection paradigm}

The tracking-by-detection paradigm was first introduced in the SORT paper~\cite{sort}. SORT combines an object detector with the KF for estimating the bounding boxes corresponding to the tracked objects and uses an association method based solely on the Intersection Over Union (IoU) measure of overlap. The object's trajectory history serves as an input to a motion model, which then predicts the object's bounding box in the current frame. A separate set of bounding boxes is obtained by passing the current frame through an object detector. These detections are then matched with the predictions to determine the tracked object's new position. Matching between the track estimations and the detections is a linear assignment problem, which can be solved using the Hungarian algorithm~\cite{linear_assignment} by employing the negative IoU as the cost measure. A match is only possible if the IoU exceeds a specific threshold, denoted as $\text{IoU}_{min}$. Finally, the state of the motion model is updated based on these detection matches. If a track is not matched with any detection for $T_{lost}$ consecutive frames, it is discarded. In this work, we use SORT as a baseline to demonstrate the effectiveness of our methodology.

\textbf{Kalman filter}. Each iteration of the KF has two phases: \textit{predict} and \textit{update}. We use $\bm{z}_i$ to denote the true state of a tracked object at time $t_i$. The measurement $\bm{x}_{i}$ is assumed to be a sample from a normal distribution $\mathcal{N}(\bm{H} \bm{z}_i, \bm{R})$ where $\bm{H}$ is a matrix that maps states to observations and $\bm{R}$ is a measurement (observation) noise covariance matrix. The prediction step estimates the mean $\hat{\bm{z}}_{i+1}$ and covariance $\hat{\bm{P}}_{i+1}$ of the Gaussian prior distribution as follows:
\begin{align}
    \hat{\bm{z}}_{i+1} &= \bm{F} \tilde{\bm{z}}_{i} \label{eq:kf_prior_mu}
    \\
    \hat{\bm{P}}_{i+1} &= \bm{F} \tilde{\bm{P}}_{i} \bm{F}^\top + \bm{Q} \label{eq:kf_prior_sigma}
\end{align}
where $\bm{F}$ is the state transition matrix defining the linear motion model, usually derived from first principles, $\bm{Q}$ is the process noise covariance matrix, and $\tilde{\bm{z}}_{i}$ and $\tilde{\bm{P}}_{i}$ are the previous KF posterior estimates. The measurement prior is also a Gaussian, with parameters
\begin{align}
    \hat{\bm{x}}_{i+1} &= \bm{H} \hat{\bm{z}}_{i+1} \label{eq:kf_prior_obs_mu}
    \\
    \hat{\bm{\Sigma}}_{i+1} &= \bm{H} \hat{\bm{P}}_{{i+1}} \bm{H}^\top + \bm{R} \label{eq:kf_prior_obs_sigma}
\end{align}
The update step estimates the Gaussian distribution of the posterior $\mathcal{N}(\tilde{z}_{i+1}, \tilde P_{i+1})$ as follows:
\begin{align}
    \bm{K}_{i+1} &= \hat{\bm{P}}_{i+1} \bm{H}^\top \hat{\bm{\Sigma}}^{-1}_{i+1} \label{eq:kf_gain}
    \\
    \Delta \bm{x}_{i+1} &= \bm{x}_{i+1} - \hat{\bm{x}}_{i+1} \label{eq:kf_innovation}
    \\
    \tilde{\bm{z}}_{i+1} &= \hat{\bm{z}}_{i+1} + \bm{K}_{i+1} \Delta \bm{x}_{i+1} \label{eq:kf_posterior_mu}
    \\
    \tilde{\bm{P}}_{i+1} &= \hat{\bm{P}}_{i+1} - \bm{K}_{i+1} \hat{\bm{\Sigma}}_{i+1} \bm{K}^\top_{i+1} \label{eq:kf_posterior_sigma}
\end{align}
where $\bm{K}_{i+1}$ is the \textit{Kalman gain}, and $\Delta \bm{x}_{i+1}$ is the \textit{innovation}. The posterior mean and covariance represent the final estimation of the object's state at time $t_{i+1}$.

In the context of the tracking-by-detection paradigm, the state vector $\bm{z}$ is typically 8-di\-mensional and consists of the horizontal and vertical pixel locations of the object's bounding box top left corner, its width $w$, height $h$, and their time derivatives. The detector provides noisy observations $\bm{x}$ of the first four states. The KF's prior estimate is used for the association between object tracks from the last frame, $t_{i}$, and detected objects in the current frame, $t_{i+1}$. Upon a successful match, the prior is combined with the matched detection using a Bayes rule to derive the posterior given by Eq.\ \eqref{eq:kf_posterior_mu} and Eq.\ \eqref{eq:kf_posterior_sigma}. During this process, the KF performs filtering of noisy and potentially corrupted measurements, i.e., detections. The process and measurement (detector) noise covariance matrices $\bm{Q}$ and $\bm{R}$ used by the KF must be adjusted. While these parameters are typically manually tuned, they can also be optimized using methods such as stochastic gradient descent to minimize the energy function in a linear Gaussian model~\cite{sarkka2013bayesian}.

\textbf{Bot-Sort Kalman filter}. The Bot-Sort~\cite{botsort} adaptive KF is used in state-of-the-art solutions for the MOT17 and MOT20 datasets, which predominantly feature linear pedestrian motion~\cite{mot20}. Bot-Sort KF uses a constant velocity linear motion model and time-dependent diagonal covariance matrices $\bm{Q}_t$ and $\bm{R}_t$, with standard deviations proportional to the dimensions (height $h$ and width $w$) of the bounding box:
\begin{equation}
\begin{split}
    \bm{Q}_t = \diag((\sigma_p \tilde{w}_{t-1})^2, (\sigma_p \tilde{h}_{t-1})^2, \\ (\sigma_p \tilde{w}_{t-1})^2, (\sigma_p \tilde{h}_{t-1})^2, \\
    (\sigma_v \tilde{w}_{t-1})^2, (\sigma_v \tilde{h}_{t-1})^2, \\ (\sigma_v \tilde{w}_{t-1})^2, (\sigma_v \tilde{h}_{t-1})^2) \label{eq:botsort_Q}
\end{split}
\end{equation}
\begin{equation}
\begin{split}
    \bm{R}_t = \diag((\sigma_m \hat{w}_{t})^2, (\sigma_m \hat{h}_{t})^2, \\ (\sigma_m \hat{w}_{t})^2, (\sigma_m \hat{h}_{t})^2) \label{eq:botsort_R}
\end{split}
\end{equation}
where $\sigma_p$, $\sigma_v$ and $\sigma_m$ are tunable parameters.

\textbf{Kalman filter disadvantages}. While the KF is highly effective, its limitation lies in modeling only linear motion. Another significant drawback, as investigated more thoroughly in~\cite{ocsort}, is the error accumulation and divergence that can occur when the update operation is skipped. This situation frequently arises in object tracking due to missed detections. In such situations, we cannot perform the update step, since the measurement $\bm{x}$ is not available, and the predicted prior is adopted as the final estimate. In this work, we show that these drawbacks can be overcome using non-linear deep learning-based motion models, which also reduce the number of domain-related design choices.

\subsection{Neural ordinary differential equations}

Neural Ordinary Differential Equations~\cite{node} (NODE) are a family of deep 
neural network models that can be understood as Residual networks 
(ResNet)~\cite{resnet} with continuous depth. A ResNet transforms a hidden state between discrete-time points $t$ 
and $t+1$ according to the following equation:
\begin{equation}
    \bm{h}_{t+1} - \bm{h}_{t} = f_{t}(\bm{h}_{t}) \label{eq:resnet_step}
\end{equation}
which can be seen as a discretization of:
\begin{equation}
  \frac{d\bm{h}(t)}{dt} = f_{t}(\bm{h}_{t}). \label{eq:resnet_to_ode}
\end{equation}
Thus, the hidden state can be modeled using an 
ODE with the initial condition $\bm{h}(0) = \bm{x}$, where $\bm{x}$ is the input data (e.g.,\ image). 
The model output is given by the final hidden state $\bm{h}(T)$ and can be computed using any black-box differential 
equation solver (e.g.\ Euler method). A Neural ODE approximates the function $f_t$ using a neural network, denoted as $f$, which includes trainable parameters $\bm\theta$:
\begin{equation}
  \frac{d\bm{h}(t)}{dt} = f(\bm{h}(t), t, \bm\theta) \label{eq:resnet_to_ode_with_theta}
\end{equation}

The loss function $L$ is applied to the outputs of the ODE solver:
\begin{align}
   L(\bm{z}(t_1)) &= L\left(\bm{z}(t_0) + \int^{t_1}_{t_0} f(\bm{z}(t), t, \bm\theta)\,dt\right) \nonumber \\
   &= L(\odesolv(\bm{z}(t_0), f, t_0, t_1, \bm\theta))\label{eq:2} \tag{2}
\end{align}
where $f$ is a neural network with parameters $\theta$, $\bm{z}(t_0)$ is the initial state at time $t_0$, and $\bm{z}(t_1)$ is the final state at time $t_1$.~\footnote{The differential equation solver $\odesolv$ usually performs multiple intermediate computation steps between time $t_0$ and $t_1$} To train this network, we calculate the gradients with respect to its parameters for every output state $\bm{z}(t)$. The main challenge in training this type of neural network lies in performing backpropagation through the ODE solver operations. To handle this issue, the ODE solver is treated as a black box, and gradients are computed using the adjoint sensitivity method~\cite{adjoint_method}. More details on this issue can be found in the original paper~\cite{node}.

The key benefits of using Neural ODEs are memory efficiency, adaptive precision of numerical integration, and the ability to naturally work with irregularly-sampled time series~\cite{odernn}. It has also been shown that these networks tend to be more robust to noise~\cite{node_robustness}. This motivates us to apply them to bounding box trajectory extrapolation in object tracking, where missing and badly aligned detections are common. 

\subsection{Conditional Neural Processes}

Conditional Neural Processes (CNP)~\cite{cnp} are a class of deep learning models that focus on the idea of learning to represent stochastic processes. CNPs operate by conditioning on a context set of observed (input, output) pairs to make predictions about unseen target points. Given a context set $\bm{C} = \{(\bm{x}_i, \bm{y}_i)\}_{i=1}^N$ representing pair of inputs $\bm{x}_i$ and corresponding observed outputs $\bm{y}_i$, and a target input set $\bm{X}_T = \{\bm{x}_j\}_{j=N+1}^M$, the CNP predicts the corresponding target outputs $\bm{Y}_T = \{\bm{y}_j\}_{j=N+1}^M$:
\begin{equation}
  \bm{y}_j \sim p(\bm{y}_j|\bm{x}_j, \bm{C}) \label{eq:cnp_modeling}
\end{equation}
where $\bm{y}_j$ is the predicted output for target point $\bm{x}_j$. The CNP model aggregates information from the context set to make predictions about each target point.

The CNP architecture is summarized by the following equations: 
\begin{align}
  \bm{r}_i &= \text{encode}(\bm{x}_i, \bm{y}_i) \label{eq:cnp_encode}
  \\
  \bm{r} &= \text{aggregate}(\{\bm{r}_i\}_{i=1}^N) \label{eq:cnp_aggregate}
  \\
  \bm{y}_j &= \text{decode}(\bm{x}_j, \bm{r}) \label{eq:cnp_decode} 
\end{align}
It involves a neural network that encodes each context point $(\bm{x}_i, \bm{y}_i)$ into a latent representation $\bm{r}_i$, as shown by Eq.\ \eqref{eq:cnp_encode}. This process is followed by an aggregation step, where these latent representations $\bm{r}_i$ are combined into a global context representation $\bm{r}$, as shown by Eq.\ \eqref{eq:cnp_aggregate}. The aggregated representation $\bm{r}$ is then used in conjunction with the observed target input $x_j$ to predict the target output $\bm{y}_j$ by another neural network, as shown by Eq.\ \eqref{eq:cnp_decode}.

CNPs are adaptable to missing context po\-ints~\cite{cnp, np, ndp}. Similar to NODEs, these attributes make CNPs well-suited for applications in the tracking-by-detection paradigm, where missing detections frequently occur. We implement a simple CNP variant---RNN-CNP, which employs an RNN (GRU) for aggregation as opposed to the standard mean aggregation used in the original CNP. This adaptation makes it more effective for object trajectory prediction.

\textbf{Attentive Conditional Neural Processes}. The Attentive Conditional Neural Process~\cite{attn_np} (ACNP) is summarized by the following equations:
\begin{align}
  \bm{R}_C &= \mhsa(\bm{X}_C, \bm{Y}_C) \label{eq:background_ancp_encode}
  \\
  \bm{r}_j &= \mhca(\bm{x}_j, \bm{X}_C, \bm{R}_C) \label{eq:background_acnp_aggregate}
  \\
  \bm{y}_j &= \text{decode}(\bm{x}_j, \bm{r}_j) \label{eq:background_acnp_decode}
\end{align}
which we explain further. ACNP enhances the standard Conditional Neural Process (CNP) by incorporating the Multi-Head Self Attention (MHSA)~\cite{attention_is_all_you_need} when computing representations as shown by Eq.\ \eqref{eq:background_ancp_encode}. This helps with modeling similarities between each of the context pairs and with achieving richer context point representations. Then, according to Eq.\ \eqref{eq:background_acnp_aggregate}, instead of using simple mean aggregation, the ACNP employs a cross-attention mechanism denoted as Multi-Head Cross Attention (MHCA)~\cite{attention_is_all_you_need}, which is designed to aggregate only the relevant context information for each target input. Within this cross-attention function, the target input $\bm{x}_j$ serves as the query, the context inputs $\bm{X}_C = \{\bm{x}_i\}^{N}_{i=1}$ act as keys, and the context representations $\bm{R}_C = \{\bm{r}_i\}^{N}_{i=1}$ function as values. In simpler terms, instead of the standard mean aggregation, the weighted mean depends on both the context and the target input. The decoding step is similar to the original CNP, but each target uses its own aggregated representation $\bm{r}_j$ instead of the global one $r$ used in CNP, as shown by Eq.\ \eqref{eq:background_acnp_decode}. The reason for this is that the attention mechanism already incorporated all of the context information into the representations of each target.

\section{Methodology}
\label{sec:methodology}

In this section we present the main methodological contributions of our work focusing on the deep learning-based filters. Since they rely on deep learning-based motion models, we describe them first and then proceed to build the filters atop of them. Finally, we propose a new method to improve the association task in tacking-by-detection paradigm of object tracking.

\subsection{Uncertainty-aware object trajectory prediction}
\label{sec:object_trajectory_prediction}

First, we define the trajectory prediction task. Assume that a trajectory is observed at time points $\bm{T}_{O} = \{t_1, t_2, t_3, \dots , t_n\}$ with corresponding observation vector values $\bm{X}_{O} = \{x_1, x_2, x_3, $\dots$ x_n\}$. These time points do not have to be equidistant. Each observation $x_i$ is given by a vector of four bounding box coordinates (top, left, width, height). 
The task is to estimate the unobserved target values $\bm{X}_{T} = \{x_{n+1}, x_{n+2}, $ $x_{n+3}, $\dots$, x_{n+m}\}$ at target time points $\bm{T}_{T} = \{t_{n+1}, t_{n+2}, \dots, t_{n+m}\}$. More concretely, we wish to model the conditional distribution of target values given the observed time points, observation values, and target time points:
\begin{equation}
  \bm{X}_{T} \sim p(\bm{X}_{T}|\bm{T}_{O}, \bm{X}_{O}, \bm{T}_{T}) \label{eq:trajectory_prediction_conditional_distribution}
\end{equation}
The motion models we propose, which model this conditional distribution, are specifically designed for the task of trajectory prediction in object tracking. They are specialized to predict dynamic movements of objects, which is crucial for accurate object tracking.

We choose to model the conditional distribution $p(\bm{X}_{T} | \bm{T}_{O}, \bm{X}_{O}, \bm{T}_{T})$ as a Gaussian distribution, where we assume conditional independence among all target variables. For simplicity, we also assume independence among the target bounding box coordinates, which allows us to model each target variable distribution with a mean vector and a diagonal covariance matrix. Put simply, each motion model produces eight values: the mean value and variance for each of the four coordinates. The assumption of independence among the target bounding box coordinates is commonly applied in trackers that employ the KF~\cite{sort, deepsort, botsort}. This not only simplifies the model but also enables more efficient computation of the loss with improved numerical stability~\cite{pytorch, gaussian_nloss}.

\textbf{Training}. Since the object trajectory targets are treated as samples from Gaussian distributions, we train our models to minimize the Gaussian negative log-likelihood (NLL) of each target point. We train all models on annotated dataset ground truths. The loss is defined as follows:
\begin{align}
    L_{nll}(\mu_{i, j}, \hat{x}_{i, j}, \hat{\sigma}_{i, j}) &= \frac{1}{2} \frac{(\mu_{i, j} - \hat{x}_{i, j})^2}{\hat{\sigma}_{i, j}^2} \nonumber \\ &+ \frac{1}{2} \log(2\pi\hat{\sigma}_{i, j}^2)  \label{eq:traj_prediction_loss}
\end{align}
where $\mu_{i, j}$ represents the $j$th ($j = 1, \ldots, 4$) target bounding box coordinate for time point $t_i$, $\hat{x}_{i, j}$ is the corresponding estimated  mean, and $\hat{\sigma}_{i, j}^2$ is the corresponding estimated variance ($j$th element on the diagonal of the diagonal covariance matrix $\hat{\bm{\Sigma}}_{i}$). The final loss is calculated by averaging across all coordinates and time points. To ensure computational stability during training, we compute the loss by imposing a small positive lower bound on $\hat{\sigma}_{i, j}$, thereby stopping it from becoming too close to zero~\cite{pytorch, gaussian_nloss}.

\subsection{Motion model architectures}
\label{sec:pmm_architectures}

We focus on three model architectures: two from the CNP family (RNN-CNP and ACNP) and one from the NODE family (RNN-ODE). During inference, each model uses a measurement buffer, which will be explained at the end of this section.

Unless specified otherwise, in our deep learning motion models all linear network layers are followed by layer normalization~\cite{layer_norm}, and then by leaky ReLU~\cite{leaky_relu}. The last layer of each proposed architecture is a linear layer without any activations or normalization, specifically adjusted for the regression task. The output of the motion model consists of 8 values, where the first 4 represent the mean values of the bounding box coordinates, and the last 4 represent the logarithms of their variances.

Instead of working with sequences of raw bounding box coordinates and time values directly, we preprocess them to extract features related to the object's motion. We define the model input features as a sequence of $t_i \parallel \bm{x}_i$, where $\bm{x}_i$ is the transformed bounding box at the $i$th trajectory step, $t_i$ is the transformed time at the $i$th trajectory step, and $\parallel$ is the concatenation operator. A detailed explanation of these transformations and their effects on the motion model performance is provided in Appendix~\ref{appendix:features}.

\textbf{AR-RNN}. As a deep motion model baseline, we consider a simple auto-regressive RNN that uses GRU as the recurrent unit. The GRU unit summarizes the trajectory features into a vector representation which is used to estimate the mean and variance of the object's position for the next step.

\textbf{RNN-ODE}. In the RNN-ODE architecture, a GRU unit within the encoder is used to summarize the observed object trajectory $(\bm{T}_O, \bm{X}_O)$ into a latent trajectory representation $\bm{z}_n$. This representation is then extrapolated in latent space using an ODE solver, deriving latent representations $\bm{z}_{n+1}, \bm{z}_{n+2}, \dots, \bm{z}_{n+m}$ for the target time points $t_{n+1}, t_{n+2}, \dots, t_{n+m}$. Finally, these latent representations are used to obtain prediction for each point. The resulting RNN-ODE architecture is summarized by the following equations:
\begin{align}
    \bm{z}_{i+1} &= \gru(\bm{z}_{i}, t_{i}\,\parallel\, \bm{x}_{i}), \nonumber \\ & \hspace*{2.25cm} i = 0,\ldots, n-1 \label{eq:rnnode_encoder} \\
    \bm{z}_{i+1} &= \odesolv(\bm{z}_{i}, \mlp_{ode}, t_{i}, t_{i+1}), \nonumber \\ & \hspace*{2.25cm} i = n,\ldots, n+m-1 \label{eq:rnnode_decoder} \\
    \hat{\bm{\mu}}_i, \hat{\bm{\Sigma}}_i &= \mlp(\bm{z}_i) \label{eq:rnnode_final}
\end{align}

\textbf{RNN-CNP}. We apply a simple but effective modification to the original CNP~\cite{cnp}.
The resulting RNN-CNP architecture is summarized by the following equations:
\begin{align}
  \bm{r}_i &= \mlp_{enc}(t_i\,\parallel\,\bm{x}_i) \label{eq:rnncnp_encode}
  \\
  \bm{r} &= \gru(\{\bm{r}_i\}_{i=1}^N) \label{eq:rnncnp_aggregate}
  \\
  \hat{\bm{\mu}}_j, \hat{\bm{\Sigma}}_j &= \mlp_{dec}(t_j\,\parallel\,\bm{r}) \label{eq:rnncnp_decode}
\end{align}
The main modification is replacing the mean aggregation with a GRU unit, which is more expressive and more suitable for sequential data. We use the time points $t_i$ combined with the bounding box features $\bm{x}_i$ for context, as can be seen in Eq.\ \eqref{eq:rnncnp_encode}. For the \textit{encode} function, we use an MLP, and for \textit{aggregate}, a GRU unit is used to obtain the context representation $\bm{r}$, as shown in Eq.\ \eqref{eq:rnncnp_aggregate}. We treat the prediction time points $t_j$ as independent queries. For the \textit{decode} step, the model estimates the target point given the query time point, based on the context $\bm{r}$, as shown in Eq.\ \eqref{eq:rnncnp_decode}. This approach offers two key benefits during decoding: it is fully parallelizable, allowing simultaneous prediction for all \textit{future} time points, and the inference time is not dependent on the temporal distance from the last observed point, a notable advantage over RNN and Neural ODE based variants. 

\textbf{ACNP}. We adapt the original ACNP~\cite{attn_np} for the trajectory prediction task by using time points as inputs and bounding box features as outputs, similar to the approach in RNN-CNP. We employ the same \textit{decode} operation as used in the RNN-CNP model, as shown in the following equations:
\begin{align}
    \bm{R}_C &= \mhsa(\bm{T}_O, \bm{X}_O) \label{eq:acnp_encode}
    \\
    \bm{r}_j &= \mhca(\bm{x}_j, \bm{X}_O, \bm{R}_C) \label{eq:acnp_aggregate}
    \\
    \hat{\bm{\mu}}_j, \hat{\bm{\Sigma}}_j &= \mlp(t_j\,\parallel\,\bm{r}_j) \label{eq:acnp_decode}
\end{align}

\textbf{Motion Model Measurements Buffer}. Having access to a detailed history of the object's movement is crucial for the motion model. This is achieved using the \textit{filter buffer}, defined by two key parameters: \textit{default\_buffer\_size} and \textit{min\_buffer\_size}. In case there are at least \textit{min\_buffer\_size} observations in the interval $[t-\text{default\_buffer\_size}, t-1]$, the observations before $t_n - \text{default\_buffer\_size}$ are ignored. Otherwise, this interval is extended so that there are always at least \textit{min\_buffer\_size} entries in the buffer. This interval extension mitigates the issue of a lack of past observation during occlusions, which generally reduces the motion model's accuracy.

The \textit{default\_buffer\_size} parameter is always ali\-gned with the motion model's history length used during training. For linear motion, the value for \textit{min\_buffer\_size} can safely be set to be equal to \textit{default\_buffer\_size}, but for non-linear motion, where dynamics change rapidly, \textit{min\_buffer\_size} value should be lower.

\subsection{Bayesian filtering with learnable motion model}
\label{sec:bayes_filters}

We propose a method that extends a non-linear probabilistic motion model in a manner analogous to the linear KF algorithm. The motion model's output serves as the Bayesian prior is combined with the likelihood of the measurements to obtain the posterior estimation. We refer to this extension as a deep Gaussian filter, as it performs filtering with the aid of deep motion models.

\begin{figure*}
  \centering
  \includegraphics[width=0.7\linewidth]{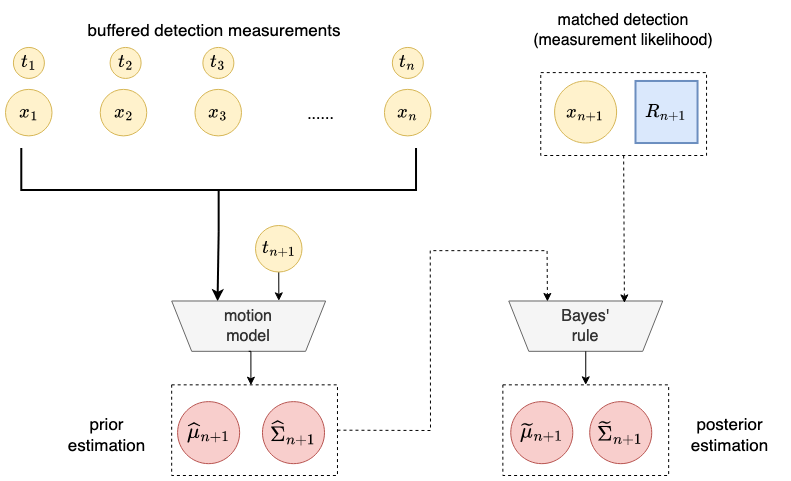}
  \caption{Visualization of a Bayesian filter incorporating non-linear, deep learning-based motion models. The measurement buffer comprises detection measurements $\{\bm{x}_1, \bm{x}_2, \dots, \bm{x}_n\}$ at times $\{t_1, t_2, \dots, t_n\}$, representing the object's trajectory history. These measurements, along with the target time $t_{n+1}$ serve as input to the motion model. The motion model predicts a prior normal distribution $\mathcal{N}(\hat{\bm{\mu}}_{n+1}, \hat{\bm{\Sigma}}_{n+1})$, which is used by the tracker in the association step. The matched detection $\bm{x}_{n+1}$, accompanied by a heuristic-based measurement noise matrix $\bm{R}_{n+1}$, represents the measurement likelihood $\mathcal{N}(\bm{x}_{n+1}, \bm{R}_{n+1})$. Finally, Bayes' rule is applied to derive the posterior estimation $\mathcal{N}(\tilde{\bm{\mu}}_{n+1}, \tilde{\bm{\Sigma}}_{n+1})$ given the prediction prior and measurement likelihood.}
  \label{fig:deep_bayes_filter}
\end{figure*}

We cannot directly substitute a deep non-linear motion model into the KF---a recursive Bayesian filter---because applying a non-linear transformation to a Gaussian-distributed random variable does not typically result in a Gaussian distribution. To address this issue, using a deep motion model we make the prediction from raw buffered measurements at each step, as shown in Figure~\ref{fig:deep_bayes_filter}. Since we lack detector measurement uncertainty, we use the same heuristic for the calculation of the measurement noise covariance matrix $\bm{R}_n$ as in Bot-Sort~\cite{botsort}. 

Our probabilistic motion models can directly estimate the prior covariance matrix $\hat{\bm{\Sigma}}_{n+1}$, thereby implicitly learning to estimate the process noise covariance matrix $\bm{Q}_{n+1}$ defined in Eq.\ ~\eqref{eq:kf_prior_sigma}. As a result, only the matrix $\bm{R}_{n+1}$ needs to be tuned. We consider the motion model's prediction as a prior distribution $\mathcal{N}(\hat{\bm{\mu}}_{n+1}, \hat{\bm{\Sigma}}_{n+1})$ for step $n+1$, with the detection $\bm{x}_{n+1}$ and the measurement noise covariance matrix $\bm{R}_{n+1}$ forming the measurement likelihood distribution $\mathcal{N}(\bm{x}_{n+1}, \bm{R}_{n+1})$. By applying Bayes' theorem to combine information from two Gaussian distributions~\cite{penny2023bayesian}, we derive the posterior distribution $\mathcal{N}(\tilde{\bm{\mu}}_{n+1}, \tilde{\bm{\Sigma}}_{n+1})$, as follows: 
\begin{align}
    \tilde{\bm{\Sigma}}_{n+1} &= (\hat{\bm{\Sigma}}^{-1}_{n+1} + \bm{R}^{-1}_{n+1})^{-1} \label{eq:bayes_rule_mu}
    \\
    \tilde{\bm{\mu}}_{n+1} &= \tilde{\bm{\Sigma}}_{n+1} \hat{\bm{\Sigma}}^{-1}_{n+1}\hat{\bm{\mu}}_{n+1} \nonumber \\
    &+ \tilde{\bm{\Sigma}}_{n+1}\bm{R}_{n+1}^{-1}\bm{x}_{n+1} \label{eq:bayes_rule_sigma}
\end{align}
This methodology is visually illustrated in Figure~\ref{fig:deep_bayes_filter}. In cases where a measurement is missing, obtaining a posterior estimation is not possible; therefore, we use the prior as a substitute.

\subsection{End-to-end learnable filtering}
\label{sec:end_to_end_filters}

When both the prior distribution and the measurement likelihood are precisely known, Bayes' rule offers a definitive method for determining the posterior distribution. However, the measurement likelihood is often not known a priori and is approximated using a heuristic based on the domain knowledge. As an alternative to such heuristic, we propose a method that directly learns to filter the measurement noise. 

\begin{figure*}
  \centering
  \includegraphics[width=0.7\linewidth]{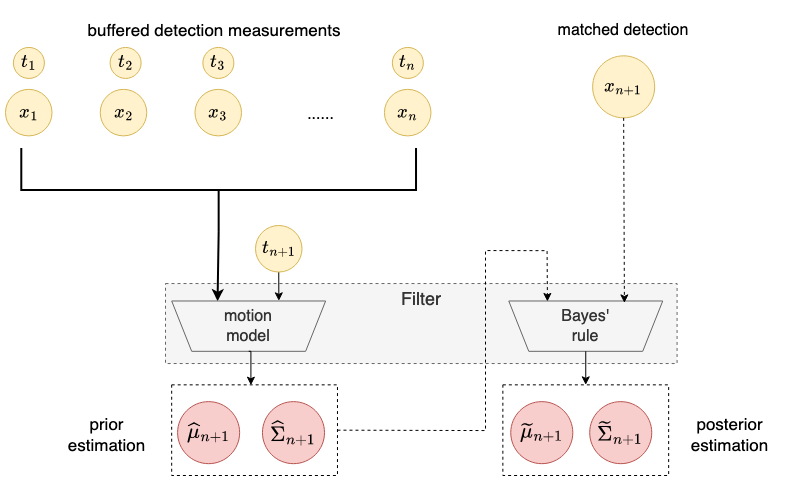}
  \caption{Overview of the two-stage end-to-end filtering process. The differences with respect to Figure~\ref{fig:deep_bayes_filter} are as follows. First, there is no requirement for providing a manually tuned measurement noise matrix $R_{n+1}$. Second, a neural network implicitly learns to filter out measurement noise, replacing the application of Bayes' rule.}
  \label{fig:deep_filter_end_to_end}
\end{figure*}

Our approach consists of two stages. In the first stage, we estimate the prior in a manner akin to conventional motion models. Then, instead of employing Bayes' theorem with a heuristically chosen covariance matrix $\bm{R}_n$, we merge the prior information with the newly observed measurement value by feeding them into  a neural network to derive the posterior estimation. All parts of this pipeline are trained together \text{end-to-end}. Hence, the name we chose for this approach. This method eliminates the need for manually tuned uncertainty hyperparameters, like the covariance matrices $\bm{Q}_n$ and $\bm{R}_n$ in the KF. For the neural network tasked with posterior estimation, we use a GRU unit aiming to learn filtering based on the measurement history. This filtering process is visualized in Figure~\ref{fig:deep_filter_end_to_end}. 

\textbf{Training}. We augment the loss by a term corresponding to the estimation of the posterior and train with ground truth annotations $\mu_{i,j}$ as targets. The new loss function is as follows:
\begin{align}
    L_{e2e}(\mu_{i, j}, \hat{\mu}_{i, j}, & \hat{\sigma}_{i, j}, \tilde{\mu}_{i, j}, \tilde{\sigma}_{i, j}) = \nonumber \\
    L_{nll}(\mu_{i, j}, \hat{\mu}_{i, j}, \hat{\sigma}_{i, j}) & + L_{nll}(\mu_{i, j}, \tilde{\mu}_{i, j}, \tilde{\sigma}_{i, j}) \label{eq:end_to_end_loss}
\end{align}
where $L_{nll}$ is defined as in Eq.\ \eqref{eq:traj_prediction_loss}. 
In real-world applications, the bounding boxes are produced by non-ideal detectors, and the measurements $\bm{x}$ can be seen as noisy versions of ground-truth bounding boxes $\bm{\mu}$. We define a denoising task by adding Gaussian noise of varying variance to $\bm{\mu}$ to synthesize the inputs $x$ to our model and train the model to reconstruct $\mu$. Thus, after training, no further hyperparameter tuning for a specific detector is required, unlike the traditional Bayesian filter. Therefore, any detector can be used with thus trained filter.

We propose two specific end-to-end filtering architectures --- NODEFilter and RNNFilter.

\begin{figure*}
  \centering
  \includegraphics[width=0.9\linewidth]{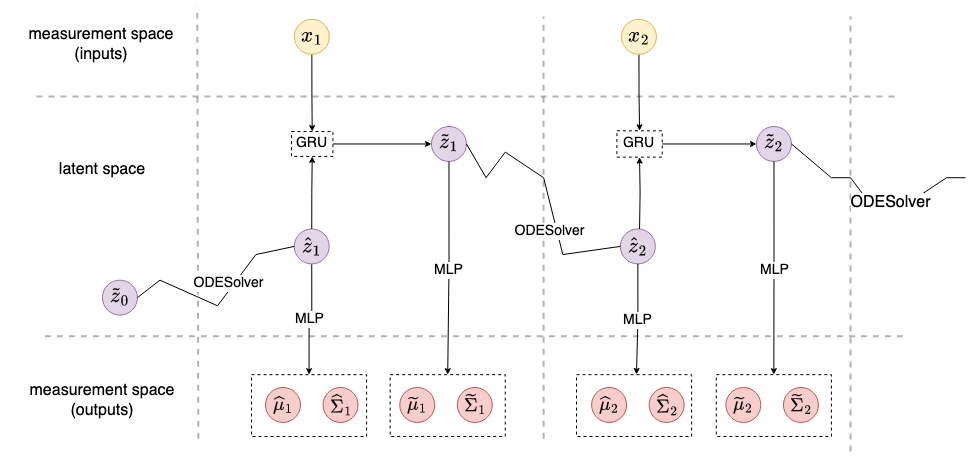}
  \caption{Visualization of the NODEFilter architecture. The latent state $\bm{z}_{n}$ summarizes the trajectory up to time $t_{n}$. Initially, $\tilde{\bm{z}}_{0}$, a zero vector, represents an empty trajectory. The ODESolver is employed to extrapolate—or \textit{predict}—the latent trajectory from $\tilde{\bm{z}}_{0}$ to $\hat{\bm{z}}_{1}$. Subsequently, the mean $\hat{\bm{\mu}}_{1}$ and covariance $\hat{\bm{\Sigma}}_{1}$ are derived from the latent representation $\hat{\bm{z}}_{1}$. Upon observing the first measurement $\bm{x}_{1}$, a GRU network produces the \textit{updated} latent state $\tilde{\bm{z}}_{1}$ by filtering out the measurement noise, resulting in final estimates of $\tilde{\bm{\mu}}_{1}$ and $\tilde{\bm{\Sigma}}_{1}$. This procedure is iteratively repeated. In scenarios where a measurement is missing, the \textit{update} step involving the GRU is omitted.}
  \label{fig:nodefilter}
\end{figure*}

\textbf{NODEFilter}. NODEFilter architecture combines NODE and GRU networks in an end-to-end fashion, resulting in high robustness at the cost of slow inference speed~\cite{node_robustness, odernn}. As seen in Figure~\ref{fig:nodefilter}, the hidden state is extrapolated by the NODE between two observations, yielding the prediction, while the GRU is used to incorporate the information from the noisy measurement and produce the final posterior estimation.
NODEFilter's state is defined through a latent vector representation, summarizing the trajectory history. The model updates the latent state with intermittent \textit{predict} and \textit{update} operations. We initialize the latent vector $\tilde{\bm{z}}_0$ to a default value (e.g., all zeros) representing an empty trajectory. The result from the \textit{predict} operation $\hat{\bm{z}}_{i+1}$ is obtained through extrapolation from the previous state using the Neural ODE with a $\mlp_{ode}$ function, as shown in Eq.\ \eqref{eq:nodefilter_ode}. The predicted latent state $\hat{\bm{z}}_{i+1}$ is combined with the new measurement $\bm{x}_{i+1}$ by the GRU neural network, resulting in the \textit{update} $\tilde{\bm{z}}_{i+1}$, as shown in Eq.\ \eqref{eq:nodefilter_gru}. Note that the latent variable is treated as the hidden state of the GRU network. The predictions are acquired by projecting from the latent space to the measurement space, as shown in Eq.\ \eqref{eq:nodefilter_mu} and Eq.\ \eqref{eq:nodefilter_sigma}. NODEFilter operations are as follows:

\begin{equation}
    \hat{\bm{z}}_{i+1} = \odesolv(\tilde{\bm{z}}_{i}, \mlp_{ode}, t_{i}, t_{i+1}) \label{eq:nodefilter_ode} 
\end{equation}
\begin{align}
    \hat{\bm{\mu}}_{i+1}, \hat{\bm{\Sigma}}_{i+1} &= \mlp_{predict}(\hat{\bm{z}}_{i+1}) \label{eq:nodefilter_mu}
    \\
    \tilde{\bm{z}}_{i+1} &= \gru(\bm{x}_{i+1}\,\parallel\,t_{i+1}, \hat{\bm{z}}_{i+1}) \label{eq:nodefilter_gru} 
    \\
    \tilde{\bm{\mu}}_{i+1}, \tilde{\bm{\Sigma}}_{i+1} &= \mlp_{update}(\tilde{\bm{z}}_{i+1}) \label{eq:nodefilter_sigma}
\end{align}

If a measurement is missing, the GRU update to the latent state is skipped, and the representation from the prediction step $\hat{\bm{z}}_{i+1}$ is used instead of $\tilde{z}_{i+1}$ to obtain the posterior estimation.

Our end-to-end filters can be used either as a recursive filter or in combination with a measurement buffer. Through our experiments, we have observed that the latter option tends to be more robust, primarily due to reduced error accumulation, and offers greater flexibility in handling trajectory input features. 

\textbf{RNNFilter}. To demonstrate the generality of our framework, we also consider the RNNFilter architecture. By substituting the continuous $\odesolv$ operation with a $\gru$ layer in Eq.~\eqref{eq:nodefilter_ode}, we develop the RNNFilter. Namely, our prediction step is as follows:
\begin{equation}
    \hat{\bm{z}}_{i+1} = \gru(t_{i+1}, \tilde{\bm{z}}_{i}) \label{eq:rnnfilter_gru} 
\end{equation}
We have deferred the exploration of filters based on neural processes to future work.

\subsection{MoveSORT}
\label{sec:movesort}

Our filters can serve as a direct replacement for the KF in any framework relying on the tracking-by-detection paradigm. We adopt the popular SORT~\cite{sort} method as a baseline, and introduce an improved association method, which we explain here. We refer to the integration of SORT with our custom filter and the new association method as MoveSORT.\footnote{The code of MoveSORT will be made public upon the end of the reviewing process.}

\textbf{Hybrid Association Method}. The standard negative IoU with gating\footnote{A match between a prior and detection is not possible if the minimum IoU constraint is not satisfied.}~\cite{bytetrack, botsort, sparsetrack, ocsort, motiontrack} does not discriminate between objects of different scales and positions. This results in a cost matrix for linear assignment based predominantly on overlap, thereby failing to adequately account for positional and scale discrepancies between the predicted and detected bounding boxes. These differences are especially important during instances of crowd occlusion, where the precision of detectors diminishes. The importance of positional and scale cues is further noticeable in scenarios involving a horizontal camera view, attributable to the perspective projection effects. For instance, the bounding box representing an individual situated further from the camera will manifest as smaller in terms of width and height compared to that of an individual located nearer. Moreover, the bounding box of the more distant individual will also be positioned upper on the projected image, even though both subjects are on the same ground level within the scene. To exploit these cues, our hybrid association method augments the conventional negative IoU technique by integrating a weighted Manhattan distance between the predicted bounding box prior $\hat{\bm{\mu}}_{n+1}$ and a detection $\bm{D}$.
\begin{align}
    C_{move}(\hat{\bm{\mu}}_{n+1}, D) &= C_{-iou}(\hat{\bm{\mu}}_{n+1}, D) \nonumber \\ &+ \lambda \cdot \|\hat{\bm{\mu}}_{n+1} - \bm{D} \|_1 \label{eq:move}
\end{align}
This proposed association method also considers positional offsets and scale. This makes association easier in crowded scenarios, where bounding boxes tend to overlap. This should lead to a lower rate of ID switches.

Although we could account for uncertainty in motion model prediction during association by using the Mahalanobis distance, which calculates the distance from the detector measurement to the motion model's estimated distribution characterized by mean $\hat{\bm{x}}_i$ and covariance $\hat{\bm{\Sigma}}_i$, we adhere to the standard practice of calculating distances between two bounding boxes as points, as demonstrated in \cite{bytetrack, botsort, sparsetrack, ocsort, bytev2}.

\section{Experimental evaluation}
\label{sec:experiments}

We evaluate the proposed methods using four MOT datasets---DanceTrack~\cite{dancetrack}, SportsMOT~\cite{sportsmot}, MOT17 and MOT20~\cite{mot20}, and one SOT dataset --- LaSOT~\cite{lasot1, lasot2}. Our interest lies in assessing the effectiveness of the proposed filters in motion-based association for Multi-Object Tracking (MOT), their accuracy in the context of Single Object Tracking (SOT), and their robustness against detector noise. We stress that we do not consider MOT17 and MOT20 proper benchmarks for our methods since they exhibit linear motion and our methods are specifically designed for non-linear motion. However, they are standard MOT benchmarks and we include them in the evaluation for transparency.

\subsection{Experimental setup}
\label{sec:implementation_details}

Here, we discuss the design choices and hyperparameters of all components of our tracker, including the detector, tracker logic, and filters.

\textbf{Detector}. We use the YOLOX detector~\cite{yolox} for all MOT datasets, with the detection confidence threshold set to $0.6$. For LaSOT, we train YOLOv8~\cite{yolov8}. Detailed information about each detector can be found in Appendix~\ref{appendix:object_detection}.

\textbf{Tracker logic}. We have determined that the choice of $\lambda = 5$ in Eq.\ \eqref{eq:move} for our hybrid association method is efficient across all datasets, ensuring that the IoU cost remains dominant, but still reducing the rate of ID switches.

A track is deleted if it is lost, i.e.\ not matched with any observations, for $T_{lost}$ consecutive frames. We tuned $T_{lost}$ separately for each dataset, choosing values of $30$, $30$, $20$, and $60$ for DanceTrack, SportsMOT, MOT17, and MOT20, respectively. 

Once an object is re-identified, linear interpolation is performed between its current bounding box position and the last known position before the track was lost, thereby filling gaps in the object track. This method is a standard approach, often used for MOT tracking~\cite{bytetrack, botsort, ocsort, deepocsort, motiontrack_byte_cmc}. For DanceTrack, linear interpolation proved to be nonbeneficial if an object is lost for more than $5$ frames. Alternative interpolation methods can be investigated in future work. A new track is initialized if an object is successfully matched for $3$ consecutive frames; otherwise, it is deleted. This approach effectively reduces the number of false positives. We set the IoU threshold $\text{IoU}_{min}$ to $0.25$ unless specified otherwise.

\textbf{Filters}. The filter's \textit{default\_buffer\_size} is always set to the length of the trajectory history used during model training. The same applies to the \textit{min\_buffer\_size} unless specified otherwise. Additional augmentations include simulated missing detections and bounding box localization errors, resulting in improved robustness to the detector errors. These augmentations are explained in detail in Appendix~\ref{appendix:augmentations}.

Trajectory features, which are explained in more detail in Appendix~\ref{appendix:features}, are preprocessed using \textit{standardized scaled first-order difference} for MOT17 and MOT20 datasets, and \textit{standardized relative to last observation} for DanceTrack, SportsMOT and LaSOT datasets. For all our Neural ODE-based architectures, we use the Runge-Kutta fourth-order method~\cite{runge_kutta} as the ODE solver, with a fixed step size of 0.25, as the default setup for both inference and training.

We use Bot-Sort implementation of the KF as a baseline for performance comparison for all datasets.

\subsection{Evaluation metrics}
\label{sec:evaluation}

In evaluating the effectiveness of our approach for multi-object tracking, we have used the higher order metrics (HOTA, AssA, DetA), IDF1, and MOTA---Multi-Object Tracking Accuracy~\cite{hota}. MOTA, determined by FN, FP, and IDs, is particularly sensitive to the detectors' performance. IDF1 indicates the quality of associations~\cite{hota}. We use these metrics for DanceTrack, SportsMOT, MOT17, and MOT20 datasets.

Moreover, we analyze the filters on the task of single-object tracking, where accuracy is measured by the average IoU between the ground truth and our prediction. We perform evaluation of both filters' prior and posterior accuracies.

\subsection{Tracker evaluation on DanceTrack}
\label{sec:dancetrack}

DanceTrack consists of 100 videos, specifically annotated for multi-object tracking of dancers. Tracking dancers, characterized by similar appearances, frequent occlusions, and fast non-linear movements, poses a significant challenge. The dataset is divided into 40 training, 25 validation, and 35 test videos~\cite{dancetrack}. 


Table~\ref{tab:results_test_dancetrack} provides a comparison of various SORT-like tracker methods that use the same detector.\footnote{Detailed information about DanceTrack's object detector setup can be found in Appendix~\ref{appendix:dancetrack_object_detection}.} To ensure a fair comparison with the meaningful baselines, all modifications that are not related to the filters or the association cost heuristic are also applied to the SORT baseline. Thus, all achieved improvements compared to the baseline are strictly the result of applying our methodology and not of using a weak baseline. Our SORT baseline, SORT-Tuned, has increased the HOTA by $3.3\%$ compared to the DanceTrack SORT tracker \cite{dancetrack}, as can be seen from Table~\ref{tab:results_test_dancetrack}. 

The only difference between MoveSORT-KF and SORT-Tuned is in the association cost heuristic. MoveSORT-KF uses our proposed hybrid loss, instead of the negative IoU cost used in SORT-Tuned. This results in an improvement of $2.1\%$ for MoveSORT-KF compared to the SORT-Tuned tracker, demonstrating the effectiveness of this simple heuristic.

Substituting the KF with any deep learning-based filter in the MoveSORT framework leads to further improvements, as can be seen in Table~\ref{tab:results_test_dancetrack}. AR-RNN proves to be a stronger baseline than KF on this dataset, leading by $0.7\%$ in HOTA. All our filters, ACNP, RNN-CNP, RNN-ODE, RNNFilter, and NODEFilter, significantly outperform the KF baseline, with NODEFilter showing the best performance, achieving a total improvement of $2.8\%$ in HOTA compared to KF, and a $2.1\%$ improvement in HOTA compared to AR-RNN. It is noteworthy that, compared to Bayesian filters, NODEFilter excels in the DetA metric, as it performs filtering more efficiently. Additionally, replacing the GRU (MoveSORT-RNNFilter) with NODE (MoveSORT-NODEFilter) in the filter extrapolation step results in a $0.3\%$ improvement in DetA performance and $0.4\%$ in MOTA. MoveSORT-NODEFilter outperforms SORT-tuned baseline by $4.9\%$, and DanceTrack SORT baseline by $8.2\%$.

\begin{table*}
\small
\centering
\begin{tabular}{l|ccccc}
Method & HOTA & DetA & AssA & MOTA & IDF1 \\
\hline
SORT~\cite{sort} & 47.9 & 72.0 & 31.2 & 91.8 & 50.8 \\
SORT-tuned & 51.2 & 80.5 & 32.6 & 91.5 & 49.7 \\
ByteTrack~\cite{bytetrack} & 47.3 & 71.6 & 31.4 & 89.5 & 52.5 \\
OC\_SORT~\cite{ocsort} & 55.1 & 80.4 & 38.0 & 89.4 & 54.9 \\
MotionTrack (DanceTrack)~\cite{motiontrack} & 52.9 & 80.9 & 34.7 & 91.3 & 53.8 \\
SparseTrack~\cite{sparsetrack} & 55.5 & 78.9 & \textbf{39.1} & 91.3 & \textbf{58.3} \\
\hline
MoveSORT-KF (ours) & 53.3 & 80.7 & 35.3 & \textit{91.7} & 51.2 \\
MoveSORT-AR-RNN (ours) & 54.0 & 80.5 & 36.3 & \textit{91.7} & 54.7 \\
MoveSORT-ACNP (ours) & 54.3 & 80.2 & 36.8 & 91.3 & 54.4 \\
MoveSORT-RNN-CNP (ours) & 55.9 & 80.6 & \textit{38.8} & \textbf{91.8} & \textit{56.5} \\
MoveSORT-RNN-ODE (ours) & 55.6 & 80.7 & 38.5 & \textbf{91.8} & 56.3 \\
MoveSORT-RNNFilter (ours) & \textit{56.0} & \textit{81.4} & 38.7 & 91.4 & 56.0 \\
MoveSORT-NODEFilter (ours) & \textbf{56.1} & \textbf{81.6} & 38.7 & \textbf{91.8} & 56.0 \\
\end{tabular}
\caption{Evaluation results for various SORT-like tracking methods on DanceTrack test set. The best results are in bold for each metric, while the second-place results are italicized.}
\label{tab:results_test_dancetrack}
\end{table*}

We also compared our best tracker, MoveSORT-NODEFilter, to other SORT-like trackers in the literature on the DanceTrack test set. Our tracker outperforms the transformer-based motion model, specifically MotionTrack~\cite{motiontrack}, by $3.2\%$ HOTA when it is trained only on the DanceTrack training dataset. MotionTrack performs stronger when trained on the SportsMOT dataset, which offers more diverse motion. However, it's important to note that we used only DanceTrack for model training. We also outperform OC\_SORT by $1.0\%$ in HOTA. The strongest SORT-like method after ours is SparseTrack~\cite{sparsetrack}, which is based on a depth estimation heuristic, but our method still surpasses SparseTrack by $0.6\%$ in terms of HOTA.

All our methods use the same post-processing algorithm,\footnote{Post-processing algorithm includes linear interpolation up to 5 frames, removal of all tracks that are shorter than 20 frames, and inclusion of track bounding boxes during the 2-frame initialization period.} which yields a consistent improvement of 0.2\% in terms of the HOTA metric. Specifically, for this dataset, we use an $IoU_{min}$ threshold of $0.2$.

\textbf{Ablation study}. We conducted an incremental ablation study to analyze the contribution of each modification from the SORT baseline to MoveSORT-NODEFilter. The results are detailed in Table~\ref{tab:ablation_study_dancetrack}. Our study leads to the following conclusions:
\begin{itemize}
    \item Increasing the $T_{lost}$ for lost tracks is crucial, especially in scenarios with frequent occlusions. The original SORT, tuned for MOT16, was effective with a $T_{lost}$ of 1. Adjusting this parameter results in a $3.2\%$ improvement in HOTA. We have chosen $T_{lost}$ of 30 (equivalent to 1.5 seconds) for this dataset.
    \item Our proposed hybrid association method further improves SORT tracker performance by $2.3\%$. This method is even more effective when combined with a more accurate filter which
    gives an improvement of $4.2\%$ in HOTA (comparison between $5$th and $7$th row in Table~\ref{tab:ablation_study_dancetrack}).
    \item Just by replacing KF with a stronger filter in SORT without changing the association method improves tracker performance by $1.5\%$ in HOTA (comparison between $2$nd and $7$th row in the Table~\ref{tab:ablation_study_dancetrack}).
\end{itemize}

\begin{table*}
    \centering
    \small
    \begin{tabular}{c|ccccc|ccccc}
        row & \makecell{$T_{lost}$} & \makecell{association \\ method} & {filter} & \makecell{$IoU_{min}$} & \makecell{post- \\ process} & {HOTA} & {DetA} & {AssA} & {MOTA} & {IDF1} \\
        \hline\hline
        1 & 1 & IoU & KF & 0.3 & No & 46.3 & 78.5 & 27.5 & 88.0 & 40.7 \\
        \hline
        2 & \cellcolor{lightgray}30 & IoU & KF & 0.3 & No & \makecell{49.5 \\ (+3.2)} & \makecell{77.8 \\ (-0.7)} & \makecell{31.6 \\ (+4.2)} & \makecell{88.9 \\ (+0.9)} & \makecell{48.4 \\ (+7.7)} \\
        \hline
        3 & 30 & \cellcolor{lightgray}IoU and L1 & KF & 0.3 & No & \makecell{51.8 \\ (+5.5)} & \makecell{77.7 \\ (-0.8)} & \makecell{34.7 \\ (+7.2)} & \makecell{89.0 \\ (+1.0)} & \makecell{51.2 \\ (+10.5)} \\
        \hline
        4 & 30 & IoU and L1 & \cellcolor{lightgray}NF & 0.3 & No & \makecell{54.1 \\ (+7.8)} & \makecell{78.5 \\ (+0.0)} & \makecell{37.5 \\ (+10.0)} & \makecell{88.9 \\ (+0.9)} & \makecell{52.9 \\ (+12.2)} \\
        \hline
        5 & 30 & IoU and L1 & NF & \cellcolor{lightgray}0.2 & No & \makecell{55.2 \\ (+8.9)} & \makecell{78.1 \\ (-0.4)} & \makecell{39.3 \\ (+11.8)} & \makecell{88.9 \\ (+0.9)} & \makecell{55.1 \\ (+14.4)} \\
        \hline
        6 & 30 & IoU and L1 & NF & 0.2 & \cellcolor{lightgray}Yes & \makecell{55.4 \\ (+9.1)} & \makecell{78.4 \\ (-0.1)} & \makecell{39.3 \\ (+11.8)} & \makecell{89.7 \\ (+1.7)} & \makecell{55.2 \\ (+14.5)} \\
        \hline\hline
        7 & 30 & \cellcolor{lightgray}IoU & NF & 0.2 & \cellcolor{lightgray}No & 51.0 & 78.0 & 33.5 & 88.8 & 50.1 \\
    \end{tabular}
    \caption{Comparison of SORT tracker performance using different parameters on the DanceTrack validation set. We use acronym NF for NODEFilter architecture. Parentheses include the total improvement compared to the SORT baseline. Shaded boxes indicate modifications made to the tracker compared to the previous row. The total improvement compared to SORT includes $9.1\%$ in terms of HOTA, $1.7\%$ in terms of MOTA, and $14.5\%$ in terms of IDF1 metrics.}
    \label{tab:ablation_study_dancetrack}
\end{table*}

\subsection{Tracker evaluation on SportsMOT}
\label{section:sportsmot}

SportsMOT is a large-scale multi-object tracking dataset comprising $240$ video sequences with over $150,000$ frames. The dataset includes three sports categories: basketball, volleyball, and football. Characterized by fast and variable-speed motion, as well as similar object appearances, SportsMOT presents a significant challenge for object tracking. The dataset is divided into training, validation, and test sets, containing $45$, $45$, and $150$ video sequences, respectively~\cite{sportsmot}.

Similar to our evaluation approach on the DanceTrack dataset, we initially configured SORT-tuned tracker to make the comparison fair. Compared to the defined default tracker parameters, we set the minimum IoU threshold $\text{IoU}_{\text{min}}$ to $0.05$. Detailed information about the used YOLOX object detector can be found in Appendix~\ref{appendix:sportsmot_object_detection}. 

Comparison of various state-of-the-art SORT-like trackers evaluated on test set is presented in Table~\ref{tab:results_test_sportsmot}. We observed that by applying our MoveSORT hybrid association method to the SORT-tuned tracker, we boosted the tracker's performance by $0.7\%$ in terms of HOTA. Further significant improvements were achieved by replacing the KF with one of our stronger filters. Our Bayesian filters (AR-RNN, ACNP, RNN-CNP, and RNN-ODE) enhanced the SORT performance by up to $3.2\%$ in terms of HOTA, and $2.0\%$ in terms of IDF1. On the other hand, our end-to-end filters (RNNFilter and NODEFilter) served as an even better replacement, yielding up to a $3.6\%$ improvement in terms of HOTA, and a dramatic improvement of $4.3\%$ in terms of IDF1.

We compare our trackers with SORT-like trackers from the literature on the SportsMOT test set. As indicated in Table~\ref{tab:results_test_sportsmot}, MoveSORT-RNNFilter outperforms OC\_SORT, the best performing SORT-like method on SportsMOT, by $0.9\%$ in terms of HOTA, $0.2\%$ in terms of MOTA, and $2.9\%$ in terms of IDF1. Noteworthy, as of the time of submitting this paper, our MoveSORT-RNN-CNP tracker is ranked first on the SportsMOT leaderboard in terms of MOTA, surpassing all other trackers.

\begin{table*}
\centering
\small
\begin{tabular}{l|ccccc}
Method & HOTA & DetA & AssA & MOTA & IDF1 \\
\hline
ByteTrack \cite{bytetrack} & 64.1 & 78.5 & 52.3 & 95.9 & 71.4 \\
OC\_SORT \cite{ocsort} & 73.7 & \textit{88.5} & 61.5 & 96.5 & 74.0 \\
SORT-Tuned & 70.3 & 86 & 57.5 & \textit{97.0} & 71.9 \\
\hline
MoveSORT-KF (ours) & 71.0 & 86.2 & 58.5 & \textbf{97.1} & 72.6 \\
MoveSORT-AR-RNN (ours) & 74.2 & \textbf{88.8} & 62.0 & \textbf{97.1} & 74.6 \\
MoveSORT-ACNP (ours) & 73.9 & \textbf{88.8} & 61.5 & \textbf{97.1} & 74.4 \\
MoveSORT-RNN-CNP (ours) & 74.2 & \textbf{88.8} & 62.0 & \textbf{97.1} & 74.6 \\
MoveSORT-RNN-ODE (ours) & 73.8 & \textbf{88.8} & 61.4 & \textbf{97.1} & 74.1 \\
MoveSORT-RNNFilter (ours)& \textbf{74.6} & 87.5 & \textbf{63.7} & 96.7 & \textbf{76.9} \\
MoveSORT-NODEFilter (ours) & \textit{74.5} & 88.0 & \textit{63.0} & 96.8 & \textit{76.2} \\
\end{tabular}
\caption{The evaluation results for various SORT-like tracking methods on the SportsMOT test set. The best results are in bold for each metric, while the second-place results are italicized. The results for the ByteTrack and OC\_SORT trackers are taken from~\cite{sportsmot}.}
\label{tab:results_test_sportsmot}
\end{table*}

\subsection{Tracker evaluation on MOT17}
\label{sec:mot17}

The MOT17 dataset consists of 14 short video scenes (7 in the training set and 7 in the test set) with non-static cameras and predominantly linear pedestrian motion. Tracker performance on MOTChallange~\cite{mot20} primarily benefits from association heuristics and to a lesser extent from motion model improvements (due to the slow-moving objects exhibiting linear motion). Therefore MOT17 and also MOT20 are not proper benchmarks for our filters which are designed for non-linear motion. However, these datasets are considered the standard for tracker evaluation and we include them in our evaluation for transparency. 

To ensure a transparent comparison of motion models, we have kept our method simple, adhering to the SORT baseline rather than adopting more complex, heavily engineered variants. We use global tracker hyperparameters and do not fine-tune them based on specific scenes in the validation or test set, arguing that this approach makes the evaluation more realistic. Additionally, to keep the design simple and better distinguish the contributions of our methodology, we do not employ Camera Motion Compensation (CMC) even though it is known to improve performance on non-static camera datasets~\cite{botsort, motiontrack_byte_cmc}.

Following the MOTChallenge\footnote{MOTChallenge (\url{https://motchallenge.net}) is a platform for evaluating the performance of multiple object tracking algorithms. It offers standardized datasets with ground truth annotations to facilitate fair and systematic comparisons across different methods.} guidelines, we submit only one tracker to the test server. A tracker is chosen based on the results on the MOT17 publicly available (training) data. In adherence to the standard approach, we divide each training scene into two halves: the first half is used for the training set, and the second half is used for the validation set~\cite{bytetrack, botsort, deepocsort, sparsetrack, motiontrack_byte_cmc}. Detailed information about the object detector setup can be found in Appendix~\ref{appendix:mot17_object_detection}.

\textbf{Method selection}. We examine all filter methods on the validation dataset, with detailed insights into the tracker's performance based on the chosen filter architecture, as shown in Table~\ref{tab:results_val_mot17}. Additionally, to show how well our new association method works, we compare SORT which uses the usual IoU association, and MoveSORT-KF which combines the IoU and L1 association methods. From the results, we can observe that the hybrid association in MoveSORT-KF improves the SORT tracker's performance by 0.47\% in HOTA and offers an 8.5\% reduction in ID switches. We can also see that all filters, except for the RNN-CNP, exhibit comparable performance. Based on the results presented in Table~\ref{tab:results_val_mot17}, we have selected MoveSORT-RNN-CNP, the tracker that uses the hybrid association method in conjunction with the Bayesian RNN-CNP filter for evaluation on the test set.

\begin{table*}
\centering
\small
\begin{tabular}{l|ccccccc}
Method & HOTA & DetA & AssA & MOTA & IDF1 & IDSW\\
\hline
SORT & 68.44 & 67.98 & 69.43 & 79.65 & 79.29 & 202 \\
\hline
MoveSORT-KF & 68.91 & \textbf{68.16} & 70.2 & 79.6 & 79.74 & \textbf{185} \\
MoveSORT-AR-RNN & 67.86 & 68.07 & 68.14 & 79.68 & 78.29 & 205 \\
MoveSORT-ACNP & 68.09 & 67.94 & 68.75 & 79.00 & 78.64 & 284 \\
MoveSORT-RNN-CNP & \textbf{69.13} & \textbf{68.19} & \textbf{70.6} & \textbf{79.78} & \textbf{80.52} & 196 \\
MoveSORT-RNN-ODE & 68.35 & 68.02 & 69.18 & 79.28 & 79.00 & 227 \\
MoveSORT-RNNFilter & 68.62 & \textbf{68.15} & 69.61 & 79.41 & 79.05 & 199 \\
MoveSORT-NODEFilter & 67.69 & 67.43 & 68.54 & 78.61 & 78.37 & 250 \\
\end{tabular}
\caption{Comparison of our trackers with the SORT baseline on the MOT17 validation set.}
\label{tab:results_val_mot17}
\end{table*}

\textbf{Evaluation}. We evaluate MoveSORT-RNN-CNP on the test set and compare it with current state-of-the-art methods that employ only motion-based association. These results are presented in Table~\ref{tab:results_test_mot17}. Our method lags behind state-of-the-art methods employing CMC by $2.9\%$ in HOTA. For methods not employing CMC, which are more comparable with our SORT-based trackers, the gap in HOTA reduces to $1.4\%$. As expected, trackers which employ advanced heuristics outperform ours which were not designed for linear motion nor apply such heuristics . However, our approach outperforms MotionTrack~\cite{motiontrack}, which is based on the SORT baseline like our method but employs a transformer-based motion model, by 2.3\% in HOTA. From the results, we conclude that our tracking method generalizes well even on small datasets.

\begin{table*}
\centering
\small
\begin{tabular}{l|cccccc}
Method & HOTA & DetA & AssA & MOTA & IDF1 & IDSW \\
\hline
MotionTrack (MOT17)~\cite{motiontrack} & 60.9 & - & 59.4 & 76.5 & 73.5 & - \\
ByteTrack$^{BT}$~\cite{bytetrack} & 63.1 & 64.5 & 62.0 & 80.3 & 77.3 & 2196 \\
BYTEv2$^{BT}$~\cite{bytev2} & 63.6 & 65.0 & 62.7 & 80.6 & 78.9 & 1239 \\
OC\_SORT~\cite{ocsort} & 63.2 & 63.2 & 63.4 & 78.0 & 77.5 & 1950 \\
MotionTrack$^{BT, *}$~\cite{motiontrack_byte_cmc} & 65.1 & 65.4 & 65.1 & 81.1 & 80.1 & 1140 \\
Bot-Sort$^{BT, *}$~\cite{botsort} & 65.0 & 64.9 & 65.5 & 80.5 & 80.2 & 1212 \\
SparseTrack$^{BT, *}$~\cite{sparsetrack} & 65.1 & 65.3 & 65.1 & 81.0 & 80.1 & 1170 \\
\hline
MoveSORT-RNN-CNP (ours) & 62.2 & 64.3 & 60.4 & 79.5 & 75.1 & 2688 \\
\end{tabular}
\caption{Evaluation results for various SORT-like state-of-the-art tracking methods on the MOT17 test set. Methods marked by (*) employ CMC and methods marked by (BT) are extensions of the ByteTrack tracker~\cite{bytetrack}.}
\label{tab:results_test_mot17}
\end{table*}

\subsection{Tracker evaluation on MOT20}
\label{sec:mot20}

The MOT20 dataset consists of eight crowded scenes with static cameras and predominantly linear pedestrian motion, with four scenes designated for the training set and the remaining four for the test set. Compared to the MOT17 dataset, the MOT20 dataset is notably larger with about 3 times more bounding boxes in total~\cite{mot20}. 

In this evaluation, we follow the MOTChallenge guidelines, as outlined in the previous section. Detailed information about the object detector setup can be found in Appendix~\ref{appendix:mot20_object_detection}.

\textbf{Model selection}. We evaluate all filters on the validation dataset and present the corresponding tracker performance in Table~\ref{tab:results_test_mot20}. Additionally, to demonstrate the efficiency of our proposed association method, we evaluate SORT with standard IoU association and MoveSORT-KF with the hybrid IoU and L1 association method. Our hybrid association in MoveSORT-KF slightly improves the SORT tracker's performance, indicated by a 0.1\% increase in HOTA and a 2\% reduction in track ID switches. We also observe that all filters, except for the AR-RNN, have comparable performance. The KF shows a slim advantage in HOTA scores, while our filters perform better in terms of MOTA. This aligns with expectations, considering the static camera and predominantly linear motion of pedestrians. Based on the results presented in Table~\ref{tab:results_val_mot20}, we have selected MoveSORT-RNN-ODE, the tracker that uses the hybrid association method in conjunction with the Bayesian RNN-ODE filter for evaluation on the test set.

\begin{table*}
\centering
\small
\begin{tabular}{l|ccccccc}
Method & HOTA & DetA & AssA & MOTA & IDF1 & IDSW \\
\hline
SORT & 61.60 & 64.36 & 59.17 & 80.54 & 77.51 & 971 \\
\hline
MoveSORT-KF & \textbf{61.7} & \textbf{64.45} & 59.28 & 80.61 & \textbf{77.67} & \textbf{954} \\
MoveSORT-AR-RNN & 58.78 & 63.22 & 54.87 & 78.65 & 72.78 & 1469 \\
MoveSORT-ACNP & 61.62 & 64.33 & 59.24 & \textbf{80.68} & 77.56 & 962 \\
MoveSORT-RNN-CNP & 61.55 & 64.38 & 59.07 & \textbf{80.71} & 77.59 & 975 \\
MoveSORT-RNN-ODE & \textbf{61.66} & 64.30 & \textbf{59.34} & \textbf{80.69} & \textbf{77.66} & 996 \\
MoveSORT-RNNFilter & 61.32 & 64.36 & 58.34 & 80.62 & 77.05 & 977 \\
MoveSORT-NODEFilter & 60.59 & 63.95 & 57.63 & 80.17 & 76.07 & 1102 \\
\end{tabular}
\caption{Evaluation results on the MOT20 validation set.}
\label{tab:results_val_mot20}
\end{table*}

\textbf{Evaluation}. We evaluate MoveSORT-RNN-ODE on the test set and compare it with current state-of-the-art methods that employ only motion-based association. These results are presented in Table~\ref{tab:results_test_mot20}. In this instance, our method lags behind others by up to 3\% in HOTA but has a relatively low number of ID switches, although we didn't specifically design it for linear movement. Still, like on MOT17, our approach outperforms MotionTrack~\cite{motiontrack}, by 2.2\% in HOTA.

\begin{table*}
\centering
\small
\begin{tabular}{l|cccccc}
Method & HOTA & DetA & AssA & MOTA & IDF1 & IDSW \\
\hline
MotionTrack (MOT20)~\cite{motiontrack} & 58.3 & - & 52.3 & 75.0 & 53.9 & - \\
ByteTrack$^{BT}$~\cite{bytetrack} & 61.3 & 63.4 & 59.6 & 77.8 & 75.2 & 1223 \\
BYTEv2$^{BT}$~\cite{bytev2} & 61.4 & 62.9 & 60.1 & 77.3 & 75.6 & 1082 \\
OC\_SORT~\cite{ocsort} & 62.4 & 62.4 & 62.5 & 75.7 & 76.3 & 942 \\
MotionTrack$^{BT, *}$~\cite{motiontrack_byte_cmc} & 62.8 & 64.0 & 61.8 & 78.0 & 76.5 & 1165 \\
Bot-Sort$^{BT, *}$~\cite{botsort} & 63.3 & 64.0 & 62.9 & 77.8 & 77.5 & 1,313 \\
SparseTrack$^{BT, *}$~\cite{sparsetrack} & 63.5 & 64.1 & 63.1 & 78.1 & 77.6 & 1120 \\
\hline
MoveSORT-RNN-ODE (ours) & 60.5 & 61.3 & 60.0 & 74.3 & 74.0 & 943 \\
\end{tabular}
\caption{Evaluation results for various SORT-Like state-of-the-art tracking methods on the MOT20 test set. Methods marked by (*) employ CMC and methods marked by (BT) are extensions of the ByteTrack tracker~\cite{bytetrack}.}
\label{tab:results_test_mot20}
\end{table*}

\subsection{Filter robustness analysis on LaSOT}
\label{sec:lasot}

LaSOT is a popular dataset specifically designed for single object tracking, containing over 1,400 video scenes recorded by non-static cameras that are evenly spread across 70 categories, amounting to more than 3.5 million frames. The extended LaSOT adds 10 new class categories to the test set not found in the initial training and validation sets, increasing the diversity for evaluation and analysis~\cite{lasot1, lasot2}. 

The dataset reflects the complex behavior of objects that irregularly disappear and reappear due to occlusions or moving out of view. Object tracking on such a dataset is not feasible without appearance features. Since we do not consider appearance features in this work, we do not perform tracker evaluation on this dataset. However, the diversity of motion patterns across numerous categories and scenes renders this dataset an excellent choice for comparing the accuracy of filters' motion models and the efficiency of their filtering approaches. We explore their robustness to detector noise, with a focus on bounding box localization errors and missed detections.

\textbf{Robustness to Gaussian noise}. We evaluate the robustness of different filters by simulating detector bounding box localization error while performing an end-to-end evaluation on 280 scenes (4 for each of 70 categories) from the LaSOT validation dataset. Gaussian noise is added to the ground truth data, with its bounding box covariance matrix being equal to $\diag((\sigma \cdot w)^2, (\sigma \cdot h)^2, (\sigma \cdot w)^2, (\sigma \cdot h)^2)$ where $\sigma$ varies from $0.0$ to $0.3$. We operate under the assumption that the noise statistics are not available for tuning any of the filters. A default measurement noise standard deviation of $0.05$ is used as a hyperparameter for all Bayesian filters. All our filters are trained using the same augmentations applied to the observed part of the trajectory. These augmentations simulate missing observations and bounding box localization errors. We also train a version of AR-RNN, denoted AR-RNN (naive), which was intentionally not subjected to these augmentations during training. We refer to the same architecture, but with augmentations applied during training as AR-RNN (robust). This approach was chosen to demonstrate the effectiveness of the applied augmentation methodology which is explained in detail in Appendix~\ref{appendix:augmentations}.

\begin{figure*}
  \centering
  \includegraphics[width=\linewidth]{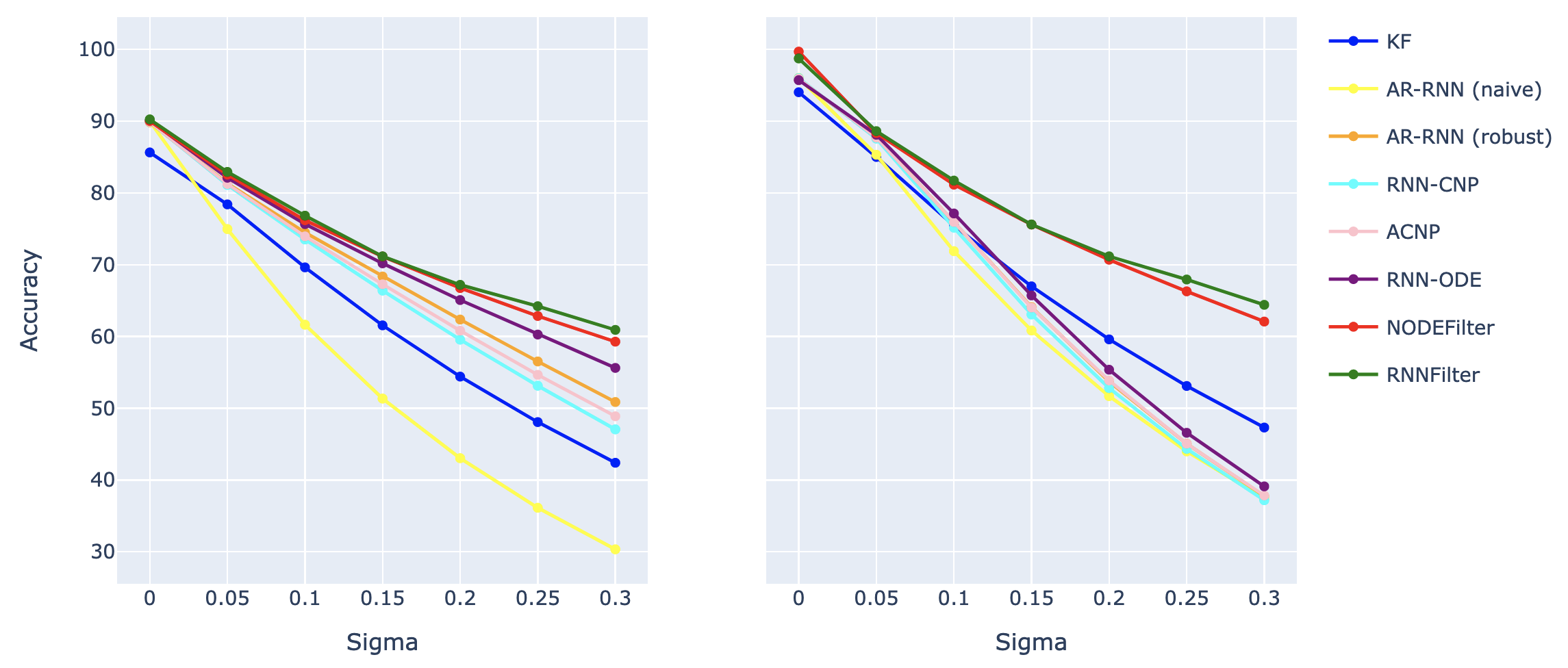}
  \caption{Evaluation of accuracy for all filters on LaSOT validation dataset for different Gaussian noise standard deviations (left---prior, right -- posterior). For exact metric values, refer to Appendix~\ref{appendix:lasot_filter_robustness}, tables~\ref{tab:lasot_gauss_prior} and~\ref{tab:lasot_gauss_posterior}.}
  \label{fig:lasot_gauss}
\end{figure*}

In Figure~\ref{fig:lasot_gauss} (left), which displays the bounding box filter prediction, it is evident that the AR-RNN (naive) significantly underperforms compared to all other methods when noise is introduced. All filters trained with augmentations surpass the performance of the KF at any noise level, including the case of noiseless detections ($\sigma=0$). Our end-to-end filters, NODEFilter and RNNFilter, demonstrate the highest robustness to Gaussian noise, as they are specifically designed for noise filtering. This difference becomes even more pronounced for posterior estimations, as seen in Figure~\ref{fig:lasot_gauss} (right). We demonstrate that our end-to-end filters can accurately filter out measurement noise at varying levels, outperforming other methods even when the true noise statistics match those assumed by the Bayesian filters (i.e.\ when noise standard deviation is $\sigma = 0.05$).

\textbf{Robustness to missed observations}. We also assess the robustness of different filters on the same set of scenes by simulating missed observations (i.e.\ false negatives). Specifically, we randomly exclude some observations and force the filters to skip the update step. We vary the probability of false negatives from 0 to 0.8. Moreover, we also add Gaussian noise to the ground truth bounding boxes to evaluate the quality of the posterior estimation. The noise standard deviation is fixed at $0.05$, aligning with the parameter value used for all Bayesian filters. Note that unlike them the NODEFilter and RNNFilter have to implicitly learn the noise statistics from partly observed trajectories. 

\begin{figure*}
  \centering
  \includegraphics[width=\linewidth]{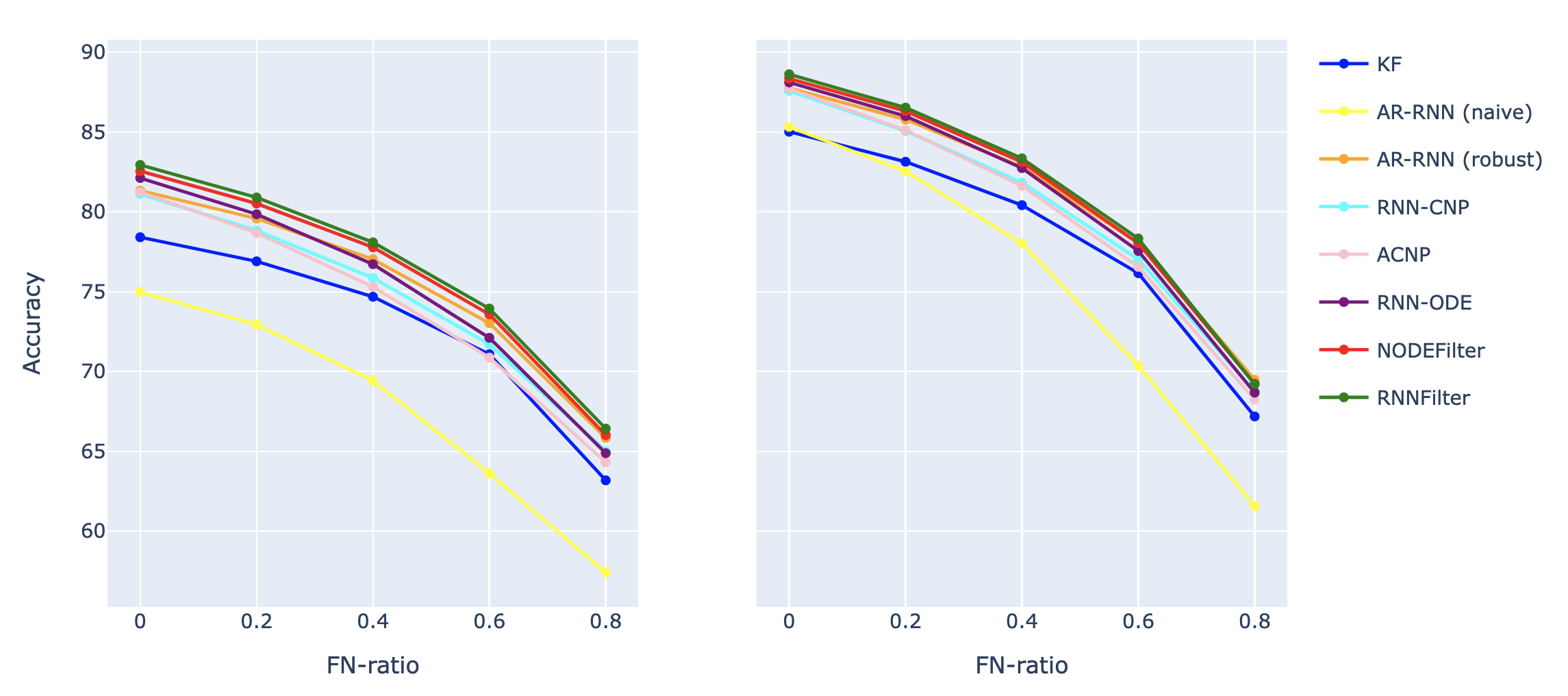}
  \caption{Evaluation of accuracy on the LaSOT validation dataset for different false negative probabilities (left---prior, right---posterior). For exact metric values, refer to Appendix~\ref{appendix:lasot_filter_robustness} tables~\ref{tab:lasot_fn_table_prior} and~\ref{tab:lasot_fn_table_posterior}.}
  \label{fig:lasot_fn}
\end{figure*}

In Figure~\ref{fig:lasot_fn} (left), we observe that the proposed augmentations significantly improve the effectiveness of AR-RNN (robust), which was trained with missing detection augmentations, compared to AR-RNN (naive), which was trained without them. KF lags considerably at different rates of missed detections. The end-to-end filters, NODEFilter and RNNFilter, deliver the strongest results. Figure~\ref{fig:lasot_fn} (right) demonstrates that the effects of missing observations on the posterior estimates are similar to those for the prior. The end-to-end filters are effective even when most of the bounding boxes are not observed.

\textbf{Robustness to real detector noise}. We also assess the robustness of filters when using a real detector. Here, localization errors and missed detections naturally occur as a consequence of the limitations of the real detector, instead of being simulated by adding noise and/or dropping the ground-truth bounding boxes, as was the case up to now. We conduct evaluations on categories where the object detector demonstrates sufficient performance for tracking-by-detection. Detailed information about the object detector setup and category selection for filter evaluation can be found in Appendix~\ref{appendix:lasot_object_detection}.

In the context of single-object tracking, altho\-ugh multiple objects may be detected in the image, only one is of interest --- the one being tracked. For this reason, we must select the most fitting detection bounding box. After performing object detection, only those bounding boxes with a confidence level higher than $0.5$ and an IoU with the ground truth of at least $0.3$ are considered. Among these candidates, the one with the highest IoU with the ground truth is chosen as the input for the filter. If no candidates meet these criteria, the detection is considered missed for that frame. Consequently, these filter inputs include both bounding box localization errors and missed detection noise, which are inherent to any real-world detector. We further discuss all the evaluation details in Appendix~\ref{appendix:robustness_to_noise_eval_details}.

From Table~\ref{tab:real_detector_lasot_prior_posterior} we observe that all filters perform similarly except for the KF which falls behind for up to $6.99\%$ in prior accuracy and $6.62\%$ in posterior accuracy. 

\begin{table*}
\centering
\small
\begin{tabular}{l|cccc}
model & prior accuracy & prior MSE & posterior accuracy & posterior MSE \\
\hline
KF & 69.03 & 6.0e-3 & 73.41 & 3.0e-3 \\
AR-RNN (naive) & 75.65 & 1.7e-3 & 79.03 & 1.6e-3 \\
AR-RNN (robust) & \textbf{76.02} & 1.7e-3 & 79.30 & \textbf{1.5e-3} \\
RNN-CNP & 75.87 & 1.7e-3 & 78.79 & \textbf{1.5e-3} \\
ACNP & 75.66 & \textbf{1.2e-3} & \textbf{79.33} & \textbf{1.5e-3} \\
RNN-ODE & 75.32 & 1.7e-3 & \textbf{79.33} & \textbf{1.5e-3} \\
NODEFilter & 75.60 & 1.7e-3 & 79.30 & \textbf{1.5e-3} \\
RNNFilter & 75.02 & 2.0e-3 & 78.58 & 1.9e-3
\end{tabular}
\caption{Results on the LaSOT validation set when the observations are produced by a real detector, instead of being synthesized from the ground truth.}
\label{tab:real_detector_lasot_prior_posterior}
\end{table*}

In Appendix~\ref{appendix:lasot_accuracy_test}, we present the results of evaluating all filters on the LaSOT test set using the ground truth bounding boxes as observations to simulate the perfect detector. In that evaluation, all data-driven filters outperform the KF by more than 10\% in terms of accuracy.

\textbf{Computational efficiency}. We measure the throughput of each method in FPS (frames per second), with a higher FPS indicating faster performance. We performed inference for each track using a single core of an Intel i7-12700K processor and a single NVIDIA GeForce RTX 3070 GPU. Results can be seen in Table~\ref{tab:filter_speed}. Neural ODE methods are known to be slower than classic RNN variants, so the RNN-ODE filter is the slowest Bayesian filter and the NODEFilter is the slowest end-to-end filter, as expected. The speed of these filters can be improved during inference at the cost of accuracy by replacing the ODE solver with a faster one (e.g.\ Runge-Kutta 4th order method with Euler method). In cases of constrained resources, the KF is easily the best option. 

\begin{table}
\centering
\small
\begin{tabular}{l|ccc}
model/FPS & predict & update & global \\
\hline
KF & \textbf{26147} & \textbf{11493} & \textbf{7984} \\
AR-RNN & 1462 & 16637 & 1397 \\
RNN-CNP & 868 & 21492 & 834 \\
ACNP & 486 & 20487 & 475 \\
RNN-ODE & 307 & 20686 & 303 \\
NODEFilter (rk4) & 53 & 1099 & 52 \\
NODEFilter (Euler) & 129 & 956 & 123 \\
RNNFilter & 85 & 1289 & 80 \\
\end{tabular}
\caption{Speed evaluation for various filters on LaSOT validation set.}
\label{tab:filter_speed}
\end{table}

\section{Related work}
\label{sec:related_work}

In this section, we compare different aspects of our approach to the relevant related work. First we discuss the  modifications we made to existing approaches to derive our motion models. Then we discuss how our filters relate to the existing ones. Finally, we compare our tracker to others that incorporate non-traditional motion models, and we discuss motion-based association heuristics related to the hybrid one we proposed.

\textbf{Motion model architectures}. In this work, we proposed different approaches to motion modeling in object tracking: RNN-CNP, ACNP, and RNN-ODE. To our knowledge, these are the first applications of Neural ODE and Neural Process architectures to object tracking. We discuss the details of the modifications we made to existing architectures to obtain these models.

As opposed to the original CNP architecture which employs mean for context aggregation~\cite{cnp, np}, the RNN-CNP we proposed uses a GRU for the same purpose. 
RNN-CNP is also related to Sequential Neural Processes (SNP)~\cite{snp}, which model a sequence of stochastic processes. Our RNN-CNP can be viewed as a special case of a conditional SNP with one context sample per frame. We use ACNP as proposed in the original ANP paper~\cite{attn_np} without significant modifications.
Our RNN-ODE consists of an efficient RNN (GRU) encoder for summarizing trajectory history, followed by the NODE decoder for trajectory extrapolation. Compared to ODE-RNN~\cite{odernn}, our encoder employs GRU only, instead of interleaved NODE and GRU steps, which makes RNN-ODE less robust~\cite{node_robustness} but more computationally efficient. The innovation that our work offers is not so much in the technical specifics of the models as in their employment as motion models.

\textbf{Filtering Methodology}. Traditional filters based on Bayesian inference are popular due to their effectiveness in various applications. Representatives of these include the Kalman Filter (KF)~\cite{sort, deepsort}, as well as its variants: the Extended Kalman Filter (EKF)~\cite{ekf}, Unscented Kalman Filter (UKF) ~\cite{ukf}, and Particle Filter (PF)~\cite{pf}. Among these, the KF is often used in multi-object tracking applications and our filtering approach is inspired by its success. Its alternatives are not frequently employed in object tracking so we did not use them in the experimental evaluation. 

Generally, KF requires a significant amount of domain knowledge to optimize its performance and guarantee satisfactory results. On the other hand, our approach is data-driven and highly adaptable to various domains, as it can learn to predict different motion patterns. We have proposed several architectures that can be effectively integrated into our non-recursive Bayesian filters, but we do not impose any constraints on architecture choices. Additionally, our Bayesian filters don't rely on hyperparameters dependent on domain knowledge except for the approximation of the measurement noise. This is unlike the Kalman Filter, which also requires initialization, a physical motion model, and process noise. Furthermore, the proposed end-to-end filter methodology eliminates the need even for the approximation of the measurement noise.

KalmanNet~\cite{kalman_net} introduced a recursive, data-driven alternative to the EKF, trained to estimate the Kalman gain using an RNN. This approach reduces the need for most KF hyperparameters while still retaining the recursive Bayesian filter pipeline. In contrast, our end-to-end filter does not necessarily require a recursive architecture and learns the distributions for both the prediction and update steps.

\textbf{Filter Model Architectures}. Based on the defined filtering methodology, we designed two new architectures---RNNFilter and NODEFilter. In comparison to the motion model architectures, these models additionally perform noise filtering on the observed measurements. Architecturally they are both related to ODE-RNN~\cite{odernn}. NODEFilter is also related to the GRU-ODE-Bayes method~\cite{gru_ode_bayes} which first used NODE in filtering. However, the differences are significant, as we elaborate in the following paragraphs.

Our proposed NODEFilter architecture closely resembles the ODE-RNN encoder~\cite{odernn}, alternating between using NODE for extrapolation between observations and GRU for observation noise filtering. In contrast, in our encoder, the NODE and GRU components are explicitly specialized for distinct sub-tasks within the architecture. While GRU is our default choice for the filtering component, our architecture allows for a flexible selection of that component. Compared to the NODEFilter architecture, the only difference in our proposed RNNFilter architecture is the usage of the RNN (GRU) instead of the NODE in the extrapolation step.

Filtering based on NODE was first proposed by the GRU-ODE-Bayes method~\cite{gru_ode_bayes} which conducts filtering by employing a continuous variant of the GRU unit, for the prior. It requires measurement likelihood uncertainty for incoming observations (measurements) to be provided from domain knowledge, analogous to the measurement noise matrix $R$ in the KF. Posterior estimation is achieved using Bayes' rule, paralleling the KF approach. In contrast, the key motivation behind our end-to-end filter methodology is eliminating the need for a measurement likelihood hyperparameter or related heuristics.

\textbf{Trackers with Non-Traditional Motion Models}. The constant velocity KF is a basic model commonly used as a baseline for comparisons and improvements. Although efficient, KF still has an issue with error accumulation during consecutive skipped update steps, often occurring due to occlusion. We discuss existing alternatives to the KF applied in object tracking and how they relate to our approach.

OC\_SORT~\cite{ocsort} addresses KF limitations by pro\-posing an observation-centric approach. Namely, instead of relying solely on the linear state estimation (i.e., an estimation-centric approach), they deal with occlusions by re-initializing the KF by interpolating missing bounding boxes after re-identification, in order to correct the error accumulation of filter parameters. The main motivation for this approach stems from the fact that detectors perform per-frame object detection independently and do not accumulate errors over time. We apply a different approach compared to OC\_SORT. Rather than attempting to rectify KF issues, we opt to replace it entirely with data-driven filters while retaining the estimation-centric approach. These filters are trained to be robust to missed detections and detector localization errors.

MotionTrack~\cite{motiontrack} introduced motion model based on the transformer architecture employing a so-called Dynamic MLP with an advanced attention mechanism. If trained on SportsMOT~\cite{sportsmot}, it shows strong results when evaluated on multiple datasets, especially compared to the KF. However, its performance drops significantly if it is not trained on SportsMOT. Another architecture, also named MotionTrack~\cite{motiontrack_byte_cmc}, employs a transformer as well but additionally incorporates an interaction module that considers the interactions between the trajectories of tracked objects. This architecture also compensates for camera motion to reduce noise from camera movement. Our tracking method does not require a specific training dataset (like SportsMOT) for strong tracking performance. Furthermore, transformer motion models do not perform filtering (i.e., they do not combine priors and detections to obtain posterior estimates).

\textbf{Motion-based association heuristics}. In this work, we proposed a hybrid association method that considers bounding box position and scale, in addition to overlap, making the tracker more effective during mutual object occlusions. This association method combines both negative IoU and L1 costs between the predicted bounding box and the candidate detection bounding box.

Compared to SORT's negative IoU association method, which is often applied~\cite{sort, bytetrack, botsort, ocsort, motiontrack}, our method is more efficient during occlusions, as the bounding box scale and position can help determine the object's depth and thus result in better matching. 

SparseTrack uses the bottom position of the bounding box for depth estimation and employs cascaded matching,\footnote{Cascaded matching in multi-object tracking is a hierarchical method in which detections are progressively matched to tracked objects using multiple criteria or stages, thereby improving tracking accuracy and robustness~\cite{deepsort, bytetrack, sparsetrack}.} starting with bounding boxes closest to the camera and iteratively incorporating those further away for association~\cite{sparsetrack}. In contrast, our method explores additional implicit cues for depth estimation, such as the width and height of the bounding box, and performs association in a single step without introducing any additional association constraints.

DeepSORT considered Mahalanobis distance between bounding boxes as it incorporated motion model uncertainty~\cite{deepsort} into the association cost. However motion model uncertainty is high during occlusion, and thus occluded objects have a higher chance of being matched with another bounding box. To address this issue, DeepSORT combined Mahalanobis distance with cascaded association, associating new detections with tracks that have more recently been successfully matched with some detection. Even though we have access to uncertainty through our filters, we intentionally refrain from using it to avoid introducing the described issues.



\section{Conclusion}
\label{sec:conclusion}

This study presents two novel filtering approaches designed to enhance object-tracking systems, particularly in non-linear motion scenarios. 
The first method is a Bayesian filter employing a trainable motion model for predicting an object's position, thereby generating a prior over the object's next bounding box. These priors are then combined with detector measurements using Bayes' rule to obtain a posterior estimation, resulting in a more precise estimate of the object's position. We propose multiple motion model architectures to be incorporated into this filtering method. These include AR-RNN, which is based solely on Recurrent Neural Networks (RNN); ACNP and RNN-CNP, both based on Neural Processes; and RNN-ODE, which is grounded in neural ordinary differential equations. We conclude that among these, the RNN-CNP is the best compromise between speed, accuracy, and robustness, offering a balanced solution in dynamic tracking environments. 

The second filtering method, an end-to-end filter, combines the motion model predictions with detector measurements by \emph{learning} to filter errors, minimizing the dependency on domain knowledge and hyperparameter adjustments in the process. We introduce two new filter architectures based on the proposed filtering methodology: RNN\-Filter, which leverages Recurrent Neural Networks (RNN), and NODEFilter, which is built upon neural ordinary differential equations. These filters are trained independently of any specific detector, making them compatible with all detectors used in the tracking-by-detection paradigm, thus eliminating the need for detector-specific tuning.

The effectiveness, robustness, and versatility of these filtering methods applied in object tracking have been thoroughly evaluated across several datasets that exhibit different motion properties. The benefits of using the proposed filters are evident in the case of datasets that exhibit non-linear motion patterns. The evaluation included different motion model architectures, demonstrating that the proposed filters achieve significant improvement compared to the Kalman Filter. Therefore, we conclude that they can replace the Kalman Filter in existing tracking systems and make a strong contribution to the multi-object tracking field. Moreover, our MoveSORT tracker, which integrates these filtering methods, includes an improved association technique that outperforms the traditional SORT method when evaluated on multiple metrics, especially in situations with complex non-linear movement and occlusions. 

Our future work, from a tracking perspective, will include the incorporation of visual appearance features, camera motion compensation, and other techniques that we did not use in this work, having aimed to focus on a proper evaluation of the proposed filters. On the other hand, we are also considering the design of more advanced end-to-end filters based on Neural Processes and their integration with Neural Ordinary Differential Equations~\cite{ndp}.

\appendix

\section{Training probabilistic motion models}
\label{appendix:training_pmm}

This section discusses input features, as described in Section~\ref{appendix:features}, the training environment detailed in Appendix~\ref{appendix:training_environment}, and the augmentations outlined in Appendix~\ref{appendix:augmentations}.

\subsection{Bounding box and temporal features}
\label{appendix:features}

A naive approach to creating input bounding box features is to use absolute coordinates. Since absolute coordinates are not translation invariant, the optimization problem for object's bounding box trajectory prediction is not trivial. Translation invariance is one of the reasons why convolutional neural networks are effective in computer vision~\cite{Goodfellow-et-al-2016}. We propose two simple transformations to convert absolute coordinates into translation invariant ones: \textit{first-order difference} and \textit{relative to last observation}.

\textit{Scaled first-order difference} is defined as $\bm{Y}_{i} = \frac{\bm{X}_{i+1} - \bm{X}_{i}}{t_{i+1} - t_{i}}, i=1, \ldots n-1$, for trajectory length of $n$. This can be seen as an approximation of the velocity at step $i$. These features are translation invariant, and also introduce a useful inductive bias, facilitating faster training. The predictions from the motion model are in the same format as the input features, necessitating the application of an inverse transformation on the model outputs to derive absolute coordinates. The inverse transform is equal to $\bm{X}_{n+i} = \bm{X}_{n} + \sum^{n+i}_{j=n+1} \bm{Y}_{j}, i=1, \ldots m$ where $m$ is the predicted trajectory length. Using the \textit{scaled first-order difference} format improves the accuracy of the RNN-CNP by 5.84\% when not standardized, and by 7.05\% when standardized, as shown in Table~\ref{tab:feature_types}.

\textit{Relative to last observation} is defined as $\bm{Y}_{i} = \bm{X}_{n} - \bm{X}_{i}, i=1, \ldots n-1$. In simple terms, the coordinate system is relative to the last observation, which can be viewed as \textit{translation invariant absolute coordinates}. The inverse transform is equal to $\bm{X}_{n+i} = \bm{X}_{n} + \bm{Y}_{n+i}, i=1, \ldots m$ where $m$ is the predicted trajectory length. Using these features improves the accuracy of RNN-CNP by 6.63\% when not standardized and by 6.95\% when standardized, as evident in Table~\ref{tab:feature_types}.

The primary advantage of using \textit{relative to last observation} features, as compared to \textit{scaled first-order differences}, lies in their robustness to noise. When the noise is high, velocity approximations can be easily disrupted, such as in cases of sudden direction changes. The \textit{relative to last observation} approach is more robust because its scale, $|\bm{X}_{n} - \bm{X}_i|$, is typically much larger than $|\frac{\bm{X}_{i+1} - \bm{X}_{i}}{t_{i+1} - t_{i}}|$, hence harder to disrupt. This is evident in Table~\ref{tab:feature_types}, where accuracy is compared under conditions of inserted bounding box Gaussian noise.

\begin{figure*}
  \centering
  \includegraphics[width=\linewidth]{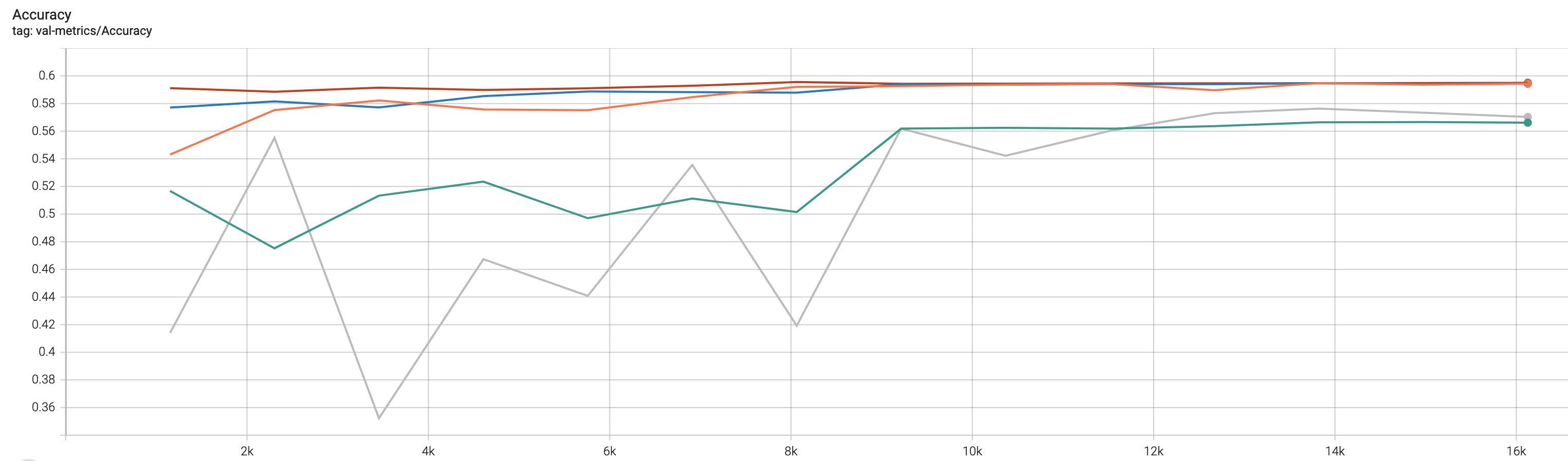}
  \caption{Comparison of the RNN-CNP motion models trained with different trajectory input features. The x-axis represents the training step, while the y-axis represents accuracy. The graph is best viewed in color: \textcolor{gray}{\textit{absolute coordinates} in gray}, \textcolor{green}{\textit{scaled first order difference} in green}, \textcolor{orange}{\textit{relative to last observation} in orange}, \textcolor{blue}{\textit{standardized scaled first order difference} in blue}, and \textcolor{red}{\textit{standardized relative to last observation} in red}.}
  \label{fig:tensorboard_features}
\end{figure*}

We can observe that effective trajectory features improve training speed, as demonstrated in Figure~\ref{fig:tensorboard_features}. Additionally, performing standardization further enhances training speed.

In addition to the features of the bounding box, we also use time information. However, since absolute time values aren't very useful, we convert them to values relative to the time point of the last observed bounding box: $T_i = t_i - t_{1}$ + 1, where $t_{1}$ is the time of the oldest observation in the measurement buffer. The final model input features are represented as $\bm{Y}_i \parallel T_i$, where $\bm{Y}_i$ are the transformed bounding box features, and $T_i$ are the transformed time features.

\begin{table*}
\centering
\small
\begin{tabular}{l|cccc}
Feature Type & \makecell{No modification} & \makecell{Gaussian\\noise} & \makecell{Missed\\observations} & \makecell{Multi-step\\(30)} \\
\hline
AC & 84.69 & 61.14 & 78.60 & 56.62\\
SFOD & 90.53 & 57.66 & 84.15 & 57.03\\
RLOC & 91.32 & 67.17 & 85.32 & 59.44\\
SSFOD & 91.74 & 59.79 & 85.78 & 59.45\\
SRLOC & 91.64 & 67.78 & 85.36 & 59.50\\
\end{tabular}
\caption{Comparison of accuracies (average overlap) of the RNN-CNP filter with different input features on the DanceTrack validation set using ground truth. We evaluate the overlap between the single-step prediction and the ground truth in three scenarios: without any modifications to the ground truth---No modification; with Gaussian noise added to the inputs---Gaussian noise ($\sigma = 0.15$); and with randomly omitted bounding boxes with probability $0.50$---Missed observations (FN ratio = $0.5$). Additionally, we consider the average overlap for multi-step predictions---Multi-step (30 steps). We use these acronyms for feature types: AC---Absolute Coordinates, SFOD---Scaled First-Order Difference, RLOC---Relative to Last Observation Coordinates, SSFOD---Standardized Scaled First-Order Difference, SRLOC---Standardized Relative to Last Observation Coordinates.}
\label{tab:feature_types}
\end{table*}

For the DanceTrack, SportsMOT, and LaSOT datasets, we found the \textit{standardized relative to last observation} feature to be the most effective. Conversely, for the MOT17 and MOT20 datasets, the \textit{standardized scaled first order difference} feature proved to be the best option. We argue that the \textit{standardized scaled first-order differences} are less effective for datasets characterized by non-linear motion due to the accumulation of errors---i.e., the sum of predicted differences---in the inverse transform required to obtain the absolute coordinates. This can be directly seen from the definition of the inverse transformation above.

\subsection{Training environment}
\label{appendix:training_environment}

All models are trained using the PyTorch framework~\cite{pytorch}. Our default setup includes a batch size of 256 and a learning rate of $0.001$, combined with a step scheduler having a multiplier of $0.1$ applied each 4 epochs. We employ the AdamW optimizer~\cite{adam, adamw} with weight decay~\cite{weight_decay}. Note that this configuration may vary between different models. For the ACNP model, we also implement a necessary warm-up, due to the use of the attention mechanism~\cite{attention_is_all_you_need}, for a period of 3 epochs.

\subsection{Augmentations}
\label{appendix:augmentations}

For our default configuration, we employ the following augmentations:
\begin{itemize}
\item \textbf{Bounding Box Gaussian Noise} — Adds Gaussian noise to the bounding boxes, enhancing the model's robustness to the detector noise. The standard deviation, denoted as \textit{sigma}, is relative to the bounding box height for the y-axis and width for the x-axis. By default, we introduce Gaussian noise augmentation. Specifically, for each trajectory, with probabilities $0.20$, $0.05$, and $0.01$ we select a standard deviation value $\sigma$ of $0.05$, $0.10$, or $0.25$, respectively and we apply noise sampled from $\mathcal{N}(0, \sigma^2)$ to that trajectory. Note that with a probability of $0.74$, there is no augmentation of this type. For end-to-end filters, we increase these probabilities to $0.60$, $0.20$, and $0.05$ respectively, as they need to learn to perform noise filtering.

\item \textbf{Remove Points} — A combination of random bounding box removal, applied with a default probability of $0.2$, and trajectory history shortening, i.e.\ clipping history from the left side, with a probability of $0.05$. The trajectory size is always maintained at the minimum of 2 points. These type of augmentations are applied at batch level.
\end{itemize}

In Table~\ref{tab:rnncnp_augmentations}, we observe that while these augmentations have no significant effect when extrapolation is performed on ground truths, they do make the model more robust to bounding box noise or missed detections.

\begin{table*}
\centering
\small
\begin{tabular}{l|cccc}
Augmentations & \makecell{Single-step} & \makecell{Gaussian\\noise} & \makecell{Missed\\observations} & \makecell{Multi-step\\(30)} \\
\hline
None & 91.79 & 53.52 & 84.10 & 59.46 \\
Gaussian noise & 91.69 & 68.10 & 84.54 & 59.53 \\
Remove points & 91.75 & 54.10 & 85.23 & 59.49 \\
Both & 91.64 & 67.78 & 85.36 & 59.50 \\
\end{tabular}
\caption{Comparison of RNN-CNP filter performance with different different augmentations. We use the same naming convention as in the Table~\ref{tab:feature_types}.}
\label{tab:rnncnp_augmentations}
\end{table*}

\section{Object detection}. 
\label{appendix:object_detection}

This section outlines the setup of the object detector for each dataset, including DanceTrack, SportsMOT, MOT17, MOT20, and LaSOT.

\subsection{DanceTrack}
\label{appendix:dancetrack_object_detection}

For the DanceTrack evaluation, we use the publicly available DanceTrack YOLOX object detector \cite{yolox}. This same detector is employed for both the validation and test sets.

\subsection{SportsMOT}
\label{appendix:sportsmot_object_detection}

For the SportsMOT test set evaluation, we use the Deep-EIoU \cite{deep_eiou} in conjunction with the YOLOX object detector \cite{yolox}. This detector was trained using both the training and validation datasets.

\subsection{MOT17}
\label{appendix:mot17_object_detection}

To address the limited data in the MOT17 dataset, the detector intended for the test set evaluation receives additional training on the validation set, in contrast to the detector employed for evaluation on the validation set, in line with the standard practice~\cite{bytetrack, botsort, deepocsort}. For tracker hyperparameter tuning and filter architecture selection on the validation dataset, we use ByteTrack's YOLOX detector, trained on CrowdHuman and the first half of each MOT17 training scene, but not on the MOT17 validation set. The second model, deployed for evaluating the test scenes, is ByteTrack's YOLOX detector trained on both the CrowdHuman and the complete MOT17 datasets. 

\subsection{MOT20}
\label{appendix:mot20_object_detection}

For evaluation on the MOT20 dataset, we employ ByteTrack's YOLOX detector trained on both CrowdHuman and the full MOT20 training set, which also includes the validation data. This detector cannot be used for tracker hyperparameter tuning and filter architecture selection on the validation dataset, due to the potential for overly optimistic results. Following the approach in \cite{deepocsort}, we employ the same detector that is used for evaluation on the MOT17 test set.

\subsection{LaSOT}
\label{appendix:lasot_object_detection}

We fine-tuned YOLOv8~\cite{yolov8} pretrained on the COCO dataset~\cite{coco} using LaSOT dataset SOT annotations. This results in a detector of lower accuracy, as SOT annotations are not necessarily optimized for multi-object detection training.~\footnote{In frames with multiple objects of interest for training a detector, only one is annotated, which is problematic for training an object detector.} For evaluating the detector, we use mAP50 (mean Average Precision for an IoU threshold of $50\%$), which is a standard metric used in YOLOv8 evaluations~\cite{yolov8}. We use only 11 classes that have a mAP50 of more than 0.4, as the noise is too extreme for object tracking with a lower mAP50 score. This set of classes includes categories with mAP50 in parentheses: flag (0.64), tank (0.58), skateboard (0.57), shark (0.41), spider (0.56), airplane (0.54), pool (0.65), tiger (0.46), bicycle (0.57), hippo (0.59), boat (0.66). Using this detector, we evaluate the accuracy of the filters' prior and posterior estimation accuracy. 

\section{Additional LaSOT results and discussion}
\label{appendix:lasot_additional}

This section contains additional information about the results and analysis on the LaSOT dataset.

\subsection{Robustness to detector noise evaluation discussion}
\label{appendix:robustness_to_noise_eval_details}

We explore the reasoning behind our evaluation setup on LaSOT, which involves real detector noise. The KF tends to accumulate errors rapidly in cases of multiple skipped update steps due to missed detections. This error is further magnified on this dataset due to the prevalence of non-linear motion, which is ill-suited to the KF's capabilities. Consequently, the predicted bounding box often drifts outside the boundaries of the frame, leading to a complete loss of tracking accuracy. In contrast, our data-driven filters adeptly learn that an object is usually positioned in the center of the camera view. To make a comparison with the KF more fair, we compute the accuracies by excluding those portions of scenes where the detection is missed for 5 or more consecutive frames.

It should be noted that this dataset lacks ground truth annotations during occlusions, meaning that motion models are not evaluated in these conditions. This scenario inadvertently favors the KF and the AR-RNN (naive) filter which are sensitive to missed detections, as can be seen in Figure~\ref{fig:lasot_fn}.

\subsection{Evaluation on test set with ground truth}
\label{appendix:lasot_accuracy_test}

We evaluate accuracy of the prediction for each filter with ground truth on the test set with 150 scenes which includes unseen categories. Based on the results in Table~\ref{tab:lasot_results}, we can observe that all non-linear data-driven filters perform similarly but drastically outperform the KF by up to $10.69\%$ in terms of accuracy.

\begin{table}
\centering
\small
\begin{tabular}{l|c}
model & prior accuracy \\
\hline
KF & 69.85 \\
AR-RNN (robust) & 80.66 \\
RNN-CNP & 80.17  \\
ACNP & 80.39 \\
RNN-ODE & 80.13 \\
NODEFilter & 80.43 \\
RNNFilter & 80.56
\end{tabular}
\caption{Evaluation of different filters on LaSOT test set in case of ground truth inputs.}
\label{tab:lasot_results}
\end{table}

\subsection{Filter robustness analysis tables}
\label{appendix:lasot_filter_robustness}

These tables include exact metric values for fi\-gures~\ref{fig:lasot_gauss} and~\ref{fig:lasot_fn}.

\begin{table*}
\centering
\small
\begin{tabular}{|l|c|c|c|c|c|c|c|}
\hline
\multicolumn{8}{|c|}{prior accuracy} \\ \hline
model/noise & 0 & 0.05 & 0.1 & 0.15 & 0.2 & 0.25 & 0.3 \\ \hline
KF & 85.65 & 78.41 & 69.63 & 61.55 & 54.41 & 48.07 & 42.4 \\
AR-RNN (naive) & 89.99 & 74.97 & 61.64 & 51.35 & 43.04 & 36.13 & 30.34 \\
AR-RNN (robust) & 89.86 & 81.33 & 74.5 & 68.36 & 62.36 & 56.52 & 50.87 \\
RNN-CNP & 89.92 & 81.12 & 73.52 & 66.4 & 59.56 & 53.14 & 47.06 \\
ACNP & 89.91 & 81.24 & 73.96 & 67.28 & 60.81 & 54.66 & 48.88 \\
RNN-ODE & 90.11 & 82.12 & 75.66 & 70.19 & 65.08 & 60.27 & 55.62 \\
NODEFilter & 90.03 & 82.55 & 76.2 & 71.15 & 66.75 & 62.86 & 59.28 \\
RNNFilter & \textbf{90.26} & \textbf{82.94} & \textbf{76.84} & \textbf{71.17} & \textbf{67.2} & \textbf{64.22} & 60.92 \\
\hline
\end{tabular}
\caption{Comparison of the performance of different filters' prior accuracy under various levels of the Gaussian noise added to ground truth on LaSOT validation set.}
\label{tab:lasot_gauss_prior}
\end{table*}

\begin{table*}
\centering
\small
\begin{tabular}{|l|c|c|c|c|c|c|c|}
\hline
\multicolumn{8}{|c|}{posterior accuracy} \\ \hline
model/noise & 0 & 0.05 & 0.1 & 0.15 & 0.2 & 0.25 & 0.3 \\ \hline
Bot-Sort KF & 94.04 & 85.02 & 75.46 & 66.99 & 59.6 & 53.1 & 47.31 \\
AR-RNN (naive) & 96.03 & 85.32 & 71.89 & 60.84 & 51.69 & 43.99 & 37.43 \\
AR-RNN (robust) & 95.82 & 87.76 & 75.94 & 64.1 & 53.73 & 45.06 & 37.76 \\
RNN-CNP & 96.01 & 87.58 & 75.17 & 63.07 & 52.81 & 44.34 & 37.19 \\
ACNP & 95.94 & 87.72 & 75.95 & 64.03 & 53.87 & 45.06 & 37.84 \\
RNN-ODE & 95.72 & 88.11 & 77.14 & 65.72 & 55.37 & 46.58 & 39.11 \\
NODEFilter & 99.7 & 88.35 & 81.16 & 75.6 & 70.7 & 66.29 & 62.08 \\
RNNFilter & \textbf{98.74} & \textbf{88.62} & \textbf{81.73} & \textbf{75.61} & \textbf{71.17} & \textbf{67.94} & \textbf{64.41} \\
\hline
\end{tabular}
\caption{Comparison of the performance of different filters' posterior accuracy under various levels of Gaussian noise applied on ground truth on LaSOT validation set.}
\label{tab:lasot_gauss_posterior}
\end{table*}

\begin{table*}
\centering
\small
\begin{tabular}{|l|c|c|c|c|c|}
\hline
 & \multicolumn{5}{c|}{prior accuracy}\\ \hline
model/FN ratio & 0 & 0.2 & 0.4 & 0.6 & 0.8 \\ \hline
Bot-Sort KF & 78.41 & 76.9 & 74.68 & 71.08 & 63.18 \\
AR-RNN (naive) & 74.97 & 72.94 & 69.42 & 63.61 & 57.41 \\
AR-RNN (robust) & 81.33 & 79.58 & 77.03 & 73.03 & 65.84\\
RNN-CNP & 81.12 & 78.83 & 75.86 & 71.71 & 65.00 \\
ACNP & 81.24 & 78.68 & 75.30 & 70.87 & 64.31 \\
RNN-ODE & 82.12 & 79.85 & 76.72 & 72.11 & 64.88 \\
NODEFilter & 82.55 & 80.52 & 77.78 & 73.58 & 66.01 \\
RNNFilter & \textbf{82.94} & \textbf{80.9} & \textbf{78.09} & \textbf{73.94} & \textbf{66.42} \\
\hline
\end{tabular}
\caption{Comparison of the performance of different filters' prior accuracies under various false negative ratios applied to ground truth on LaSOT validation set.}

\label{tab:lasot_fn_table_prior}
\end{table*}

\begin{table*}
\centering
\small
\begin{tabular}{|l|c|c|c|c|c|}
\hline
 & \multicolumn{5}{c|}{posterior accuracy} \\ \hline
model/FN ratio & 0 & 0.2 & 0.4 & 0.6 & 0.8 \\ \hline
Bot-Sort KF & 85.02 & 83.14 & 80.42 & 76.16 & 67.18 \\
AR-RNN (naive) & 85.32 & 82.54 & 78.02 & 70.37 & 61.57 \\
AR-RNN (robust) & 87.76 & 85.78 & 82.83 & \textbf{78.04} & \textbf{69.48} \\
RNN-CNP & 87.58 & 85.07 & 81.84 & 77.02 & 68.76 \\
ACNP & 87.72 & 85.12 & 81.62 & 76.55 & 68.24 \\
RNN-ODE & 88.11 & 86.00 & 82.74 & 77.55 & 68.67 \\
NODEFilter & 88.35 & 86.31 & 83.12 & 78.04 & 69.22 \\
RNNFilter & \textbf{88.62} & \textbf{86.53} & \textbf{83.35} & \textbf{78.33} & \textbf{69.21} \\
\hline
\end{tabular}
\caption{Comparison of the performance of different filters' posterior accuracies under various false negative ratios applied to ground truth on LaSOT validation set.}

\label{tab:lasot_fn_table_posterior}
\end{table*}

\clearpage

\bibliography{main}


\begin{thebibliography}{55}
\ifx \bisbn   \undefined \def \bisbn  #1{ISBN #1}\fi
\ifx \binits  \undefined \def \binits#1{#1}\fi
\ifx \bauthor  \undefined \def \bauthor#1{#1}\fi
\ifx \batitle  \undefined \def \batitle#1{#1}\fi
\ifx \bjtitle  \undefined \def \bjtitle#1{#1}\fi
\ifx \bvolume  \undefined \def \bvolume#1{\textbf{#1}}\fi
\ifx \byear  \undefined \def \byear#1{#1}\fi
\ifx \bissue  \undefined \def \bissue#1{#1}\fi
\ifx \bfpage  \undefined \def \bfpage#1{#1}\fi
\ifx \blpage  \undefined \def \blpage #1{#1}\fi
\ifx \burl  \undefined \def \burl#1{\textsf{#1}}\fi
\ifx \doiurl  \undefined \def \doiurl#1{\url{https://doi.org/#1}}\fi
\ifx \betal  \undefined \def \betal{\textit{et al.}}\fi
\ifx \binstitute  \undefined \def \binstitute#1{#1}\fi
\ifx \binstitutionaled  \undefined \def \binstitutionaled#1{#1}\fi
\ifx \bctitle  \undefined \def \bctitle#1{#1}\fi
\ifx \beditor  \undefined \def \beditor#1{#1}\fi
\ifx \bpublisher  \undefined \def \bpublisher#1{#1}\fi
\ifx \bbtitle  \undefined \def \bbtitle#1{#1}\fi
\ifx \bedition  \undefined \def \bedition#1{#1}\fi
\ifx \bseriesno  \undefined \def \bseriesno#1{#1}\fi
\ifx \blocation  \undefined \def \blocation#1{#1}\fi
\ifx \bsertitle  \undefined \def \bsertitle#1{#1}\fi
\ifx \bsnm \undefined \def \bsnm#1{#1}\fi
\ifx \bsuffix \undefined \def \bsuffix#1{#1}\fi
\ifx \bparticle \undefined \def \bparticle#1{#1}\fi
\ifx \barticle \undefined \def \barticle#1{#1}\fi
\bibcommenthead
\ifx \bconfdate \undefined \def \bconfdate #1{#1}\fi
\ifx \botherref \undefined \def \botherref #1{#1}\fi
\ifx \url \undefined \def \url#1{\textsf{#1}}\fi
\ifx \bchapter \undefined \def \bchapter#1{#1}\fi
\ifx \bbook \undefined \def \bbook#1{#1}\fi
\ifx \bcomment \undefined \def \bcomment#1{#1}\fi
\ifx \oauthor \undefined \def \oauthor#1{#1}\fi
\ifx \citeauthoryear \undefined \def \citeauthoryear#1{#1}\fi
\ifx \endbibitem  \undefined \def \endbibitem {}\fi
\ifx \bconflocation  \undefined \def \bconflocation#1{#1}\fi
\ifx \arxivurl  \undefined \def \arxivurl#1{\textsf{#1}}\fi
\csname PreBibitemsHook\endcsname

\bibitem[\protect\citeauthoryear{Balasubramaniam and Pasricha}{2022}]{object_tracking_in_autunomous_driving}
\begin{bchapter}
\bauthor{\bsnm{Balasubramaniam}, \binits{A.}},
\bauthor{\bsnm{Pasricha}, \binits{S.}}:
\bctitle{Object detection in autonomous vehicles: Status and open challenges}.
In: \bbtitle{arXiv Preprint}
(\byear{2022}).
\bcomment{Arxiv: 2201.07706}
\end{bchapter}
\endbibitem

\bibitem[\protect\citeauthoryear{Caesar et~al.}{2020}]{nuScenes}
\begin{bchapter}
\bauthor{\bsnm{Caesar}, \binits{H.}},
\bauthor{\bsnm{Bankiti}, \binits{V.}},
\bauthor{\bsnm{Lang}, \binits{A.H.}},
\bauthor{\bsnm{Vora}, \binits{S.}},
\bauthor{\bsnm{Liong}, \binits{V.E.}},
\bauthor{\bsnm{Xu}, \binits{Q.}},
\bauthor{\bsnm{Krishnan}, \binits{A.}},
\bauthor{\bsnm{Pan}, \binits{Y.}},
\bauthor{\bsnm{Baldan}, \binits{G.}},
\bauthor{\bsnm{Beijbom}, \binits{O.}}:
\bctitle{nuscenes: A multimodal dataset for autonomous driving}.
In: \bbtitle{CVPR}
(\byear{2020}).
\bcomment{Arxiv: 1903.11027}
\end{bchapter}
\endbibitem

\bibitem[\protect\citeauthoryear{Xu et~al.}{2023}]{object_tracking_in_robotics}
\begin{bchapter}
\bauthor{\bsnm{Xu}, \binits{Z.}},
\bauthor{\bsnm{Zhan}, \binits{X.}},
\bauthor{\bsnm{Xiu}, \binits{Y.}},
\bauthor{\bsnm{Suzuki}, \binits{C.}},
\bauthor{\bsnm{Shimada}, \binits{K.}}:
\bctitle{Onboard dynamic-object detection and tracking for autonomous robot navigation with rgb-d camera}.
In: \bbtitle{IEEE Robotics and Automation Letters}
(\byear{2023}).
\bcomment{Arxiv: 2303.00132}
\end{bchapter}
\endbibitem

\bibitem[\protect\citeauthoryear{Urbann et~al.}{2021}]{object_tracking_in_surveillence}
\begin{bchapter}
\bauthor{\bsnm{Urbann}, \binits{O.}},
\bauthor{\bsnm{Bredtmann}, \binits{O.}},
\bauthor{\bsnm{Otten}, \binits{M.}},
\bauthor{\bsnm{Richter}, \binits{J.-P.}},
\bauthor{\bsnm{Bauer}, \binits{T.}},
\bauthor{\bsnm{Zibriczky}, \binits{D.}}:
\bctitle{Online and real-time tracking in a surveillance scenario}.
In: \bbtitle{5th Workshop on Long-term Human Motion Prediction}
(\byear{2021}).
\bcomment{Arxiv: 2106.01153}
\end{bchapter}
\endbibitem

\bibitem[\protect\citeauthoryear{Ge et~al.}{2021}]{yolox}
\begin{bchapter}
\bauthor{\bsnm{Ge}, \binits{Z.}},
\bauthor{\bsnm{Liu}, \binits{S.}},
\bauthor{\bsnm{Wang}, \binits{F.}},
\bauthor{\bsnm{Li}, \binits{Z.}},
\bauthor{\bsnm{Sun}, \binits{J.}}:
\bctitle{Yolox: Exceeding yolo series in 2021}.
In: \bbtitle{arXiv Preprint}
(\byear{2021}).
\bcomment{Arxiv: 2107.08430v2}
\end{bchapter}
\endbibitem

\bibitem[\protect\citeauthoryear{Glenn et~al.}{2023}]{yolov8}
\begin{botherref}
\oauthor{\bsnm{Glenn}, \binits{J.}},
\oauthor{\bsnm{Ayush}, \binits{C.}},
\oauthor{\bsnm{Jing}, \binits{Q.}}:
YOLO by Ultralytics
(2023).
\url{https://github.com/ultralytics/ultralytics}
\end{botherref}
\endbibitem

\bibitem[\protect\citeauthoryear{Ren et~al.}{2015}]{faster_rcnn}
\begin{bchapter}
\bauthor{\bsnm{Ren}, \binits{S.}},
\bauthor{\bsnm{He}, \binits{K.}},
\bauthor{\bsnm{Girshick}, \binits{R.}},
\bauthor{\bsnm{Sun}, \binits{J.}}:
\bctitle{Faster r-cnn: Towards real-time object detection with region proposal networks}.
In: \bbtitle{NIPS}
(\byear{2015}).
\bcomment{Arxiv: 1506.01497}
\end{bchapter}
\endbibitem

\bibitem[\protect\citeauthoryear{Carion et~al.}{2020}]{detr}
\begin{bchapter}
\bauthor{\bsnm{Carion}, \binits{N.}},
\bauthor{\bsnm{Massa}, \binits{F.}},
\bauthor{\bsnm{Synnaeve}, \binits{G.}},
\bauthor{\bsnm{Usunier}, \binits{N.}},
\bauthor{\bsnm{Kirillov}, \binits{A.}},
\bauthor{\bsnm{Zagoruyko}, \binits{S.}}:
\bctitle{End-to-end object detection with transformers}.
In: \bbtitle{ECCV}
(\byear{2020}).
\bcomment{Arxiv: 2005.12872}
\end{bchapter}
\endbibitem

\bibitem[\protect\citeauthoryear{Bewley et~al.}{2016}]{sort}
\begin{bchapter}
\bauthor{\bsnm{Bewley}, \binits{A.}},
\bauthor{\bsnm{Ge}, \binits{Z.}},
\bauthor{\bsnm{Ott}, \binits{L.}},
\bauthor{\bsnm{Ramos}, \binits{F.}},
\bauthor{\bsnm{Upcroft}, \binits{B.}}:
\bctitle{Simple online and realtime tracking}.
In: \bbtitle{ICIP}
(\byear{2016}).
\bcomment{Arxiv: 1602.00763}
\end{bchapter}
\endbibitem

\bibitem[\protect\citeauthoryear{Wojke et~al.}{2017}]{deepsort}
\begin{bchapter}
\bauthor{\bsnm{Wojke}, \binits{N.}},
\bauthor{\bsnm{Bewley}, \binits{A.}},
\bauthor{\bsnm{Paulus}, \binits{D.}}:
\bctitle{Simple online and realtime tracking with a deep association metric}.
In: \bbtitle{ICIP}
(\byear{2017}).
\bcomment{Arxiv: 1703.07402}
\end{bchapter}
\endbibitem

\bibitem[\protect\citeauthoryear{Wang et~al.}{2020}]{jde}
\begin{bchapter}
\bauthor{\bsnm{Wang}, \binits{Z.}},
\bauthor{\bsnm{Zheng}, \binits{L.}},
\bauthor{\bsnm{Liu}, \binits{Y.}},
\bauthor{\bsnm{Li}, \binits{Y.}},
\bauthor{\bsnm{Wang}, \binits{S.}}:
\bctitle{Towards real-time multi-object tracking}.
In: \bbtitle{ECCV}
(\byear{2020}).
\bcomment{Arxiv: 1909.12605}
\end{bchapter}
\endbibitem

\bibitem[\protect\citeauthoryear{Aharon et~al.}{2022}]{botsort}
\begin{bchapter}
\bauthor{\bsnm{Aharon}, \binits{N.}},
\bauthor{\bsnm{Orfaig}, \binits{R.}},
\bauthor{\bsnm{Bobrovsky}, \binits{B.-Z.}}:
\bctitle{Bot-sort: Robust associations multi-pedestrian tracking}.
In: \bbtitle{arXiv Preprint}
(\byear{2022}).
\bcomment{Arxiv: 2206.14651}
\end{bchapter}
\endbibitem

\bibitem[\protect\citeauthoryear{Dendorfer et~al.}{2020}]{mot20}
\begin{bchapter}
\bauthor{\bsnm{Dendorfer}, \binits{P.}},
\bauthor{\bsnm{Rezatofighi}, \binits{H.}},
\bauthor{\bsnm{Milan}, \binits{A.}},
\bauthor{\bsnm{Shi}, \binits{J.}},
\bauthor{\bsnm{Cremers}, \binits{D.}},
\bauthor{\bsnm{Reid}, \binits{I.}},
\bauthor{\bsnm{Roth}, \binits{S.}},
\bauthor{\bsnm{Schindler}, \binits{K.}},
\bauthor{\bsnm{Leal-Taixe}, \binits{L.}}:
\bctitle{Mot20: A benchmark for multi object tracking in crowded scenes}.
In: \bbtitle{arXiv Preprint}
(\byear{2020}).
\bcomment{Arxiv: 2003.09003}
\end{bchapter}
\endbibitem

\bibitem[\protect\citeauthoryear{Sun et~al.}{2021}]{dancetrack}
\begin{bchapter}
\bauthor{\bsnm{Sun}, \binits{P.}},
\bauthor{\bsnm{Cao}, \binits{J.}},
\bauthor{\bsnm{Jiang}, \binits{Y.}},
\bauthor{\bsnm{Yuan}, \binits{Z.}},
\bauthor{\bsnm{Bai}, \binits{S.}},
\bauthor{\bsnm{Kitani}, \binits{K.}},
\bauthor{\bsnm{Luo}, \binits{P.}}:
\bctitle{Dancetrack: Multi-object tracking in uniform appearance and diverse motion}.
In: \bbtitle{CVPR}
(\byear{2021}).
\bcomment{Arxiv: 2111.14690}
\end{bchapter}
\endbibitem

\bibitem[\protect\citeauthoryear{Gordon et~al.}{1993}]{pf}
\begin{bchapter}
\bauthor{\bsnm{Gordon}, \binits{N.J.}},
\bauthor{\bsnm{Salmond}, \binits{D.J.}},
\bauthor{\bsnm{Smith}, \binits{A.F.M.}}:
\bctitle{Novel approach to nonlinear/non-gaussian bayesian state estimation}.
In: \bbtitle{International Society for Optics and Photonics}
(\byear{1993})
\end{bchapter}
\endbibitem

\bibitem[\protect\citeauthoryear{Gruber}{1967}]{ekf}
\begin{bchapter}
\bauthor{\bsnm{Gruber}, \binits{M.}}:
\bctitle{An approach to target tracking}.
In: \bbtitle{MIT Lexington Lincoln Lab}
(\byear{1967})
\end{bchapter}
\endbibitem

\bibitem[\protect\citeauthoryear{Julier and Uhlmann}{1997}]{ukf}
\begin{bchapter}
\bauthor{\bsnm{Julier}, \binits{S.J.}},
\bauthor{\bsnm{Uhlmann}, \binits{J.K.}}:
\bctitle{New extension of the kalman filter to nonlinear systems}.
In: \bbtitle{International Society for Optics and Photonics}
(\byear{1997})
\end{bchapter}
\endbibitem

\bibitem[\protect\citeauthoryear{Chen et~al.}{2018}]{node}
\begin{bchapter}
\bauthor{\bsnm{Chen}, \binits{R.T.Q.}},
\bauthor{\bsnm{Rubanova}, \binits{Y.}},
\bauthor{\bsnm{Bettencourt}, \binits{J.}},
\bauthor{\bsnm{Duvenaud}, \binits{D.}}:
\bctitle{Neural ordinary differential equations}.
In: \bbtitle{NIPS}
(\byear{2018}).
\bcomment{Arxiv: 1806.07366}
\end{bchapter}
\endbibitem

\bibitem[\protect\citeauthoryear{Garnelo et~al.}{2018a}]{np}
\begin{bchapter}
\bauthor{\bsnm{Garnelo}, \binits{M.}},
\bauthor{\bsnm{Schwarz}, \binits{J.}},
\bauthor{\bsnm{Rosenbaum}, \binits{D.}},
\bauthor{\bsnm{Viola}, \binits{F.}},
\bauthor{\bsnm{Rezende}, \binits{D.J.}},
\bauthor{\bsnm{Eslami}, \binits{S.M.A.}},
\bauthor{\bsnm{Teh}, \binits{Y.W.}}:
\bctitle{Neural processes}.
In: \bbtitle{ICML}
(\byear{2018}).
\bcomment{Arxiv: 1807.01622}
\end{bchapter}
\endbibitem

\bibitem[\protect\citeauthoryear{Garnelo et~al.}{2018b}]{cnp}
\begin{bchapter}
\bauthor{\bsnm{Garnelo}, \binits{M.}},
\bauthor{\bsnm{Rosenbaum}, \binits{D.}},
\bauthor{\bsnm{Maddison}, \binits{C.J.}},
\bauthor{\bsnm{Tiago~Ramalho}, \binits{D.S.}},
\bauthor{\bsnm{Shanahan}, \binits{M.}},
\bauthor{\bsnm{Teh}, \binits{Y.W.}},
\bauthor{\bsnm{Danilo J.~Rezende}, \binits{S.M.A.E.}}:
\bctitle{Conditional neural processes}.
In: \bbtitle{The Proceedings of Machine Learning Research}
(\byear{2018}).
\bcomment{Arxiv: 1807.01613}
\end{bchapter}
\endbibitem

\bibitem[\protect\citeauthoryear{Zhang et~al.}{2022}]{bytetrack}
\begin{bchapter}
\bauthor{\bsnm{Zhang}, \binits{Y.}},
\bauthor{\bsnm{Sun}, \binits{P.}},
\bauthor{\bsnm{Jiang}, \binits{Y.}},
\bauthor{\bsnm{Yu}, \binits{D.}},
\bauthor{\bsnm{Weng}, \binits{F.}},
\bauthor{\bsnm{Yuan}, \binits{Z.}},
\bauthor{\bsnm{Luo}, \binits{P.}},
\bauthor{\bsnm{Liu}, \binits{W.}},
\bauthor{\bsnm{Wang}, \binits{X.}}:
\bctitle{Bytetrack: Multi-object tracking by associating every detection box}.
In: \bbtitle{ECCV}
(\byear{2022}).
\bcomment{Arxiv: 2110.06864}
\end{bchapter}
\endbibitem

\bibitem[\protect\citeauthoryear{Ramshaw et~al.}{2012}]{linear_assignment}
\begin{bchapter}
\bauthor{\bsnm{Ramshaw}, \binits{L.}},
\bauthor{\bsnm{E}, \binits{R.}},
\bauthor{\bsnm{Tarjan}}:
\bctitle{On minimum-cost assignments in unbalanced bipartite graphs}.
In: \bbtitle{HP Laboratories}
(\byear{2012})
\end{bchapter}
\endbibitem

\bibitem[\protect\citeauthoryear{S{\"a}rkk{\"a}}{2013}]{sarkka2013bayesian}
\begin{bbook}
\bauthor{\bsnm{S{\"a}rkk{\"a}}, \binits{S.}}:
\bbtitle{Bayesian Filtering and Smoothing}.
\bpublisher{Cambridge University Press}, \blocation{???}
(\byear{2013})
\end{bbook}
\endbibitem

\bibitem[\protect\citeauthoryear{Cao et~al.}{2022}]{ocsort}
\begin{bchapter}
\bauthor{\bsnm{Cao}, \binits{J.}},
\bauthor{\bsnm{Pang}, \binits{J.}},
\bauthor{\bsnm{Weng}, \binits{X.}},
\bauthor{\bsnm{Khirodkar}, \binits{R.}},
\bauthor{\bsnm{Kitani}, \binits{K.}}:
\bctitle{Observation-centric sort: Rethinking sort for robust multi-object tracking}.
In: \bbtitle{CVPR}
(\byear{2022}).
\bcomment{Arxiv: 2203.14360}
\end{bchapter}
\endbibitem

\bibitem[\protect\citeauthoryear{He et~al.}{2016}]{resnet}
\begin{bchapter}
\bauthor{\bsnm{He}, \binits{K.}},
\bauthor{\bsnm{Zhang}, \binits{X.}},
\bauthor{\bsnm{Ren}, \binits{S.}},
\bauthor{\bsnm{Sun}, \binits{J.}}:
\bctitle{Deep residual learning for image recognition}.
In: \bbtitle{CVPR}
(\byear{2016}).
\bcomment{Arxiv: 1512.03385}
\end{bchapter}
\endbibitem

\bibitem[\protect\citeauthoryear{Pontryagin et~al.}{1962}]{adjoint_method}
\begin{bbook}
\bauthor{\bsnm{Pontryagin}, \binits{L.S.}},
\bauthor{\bsnm{Mishchenko}, \binits{E.}},
\bauthor{\bsnm{Boltyanskii}, \binits{V.}},
\bauthor{\bsnm{Gamkrelidze}, \binits{R.}}:
\bbtitle{The Mathematical Theory of Optimal Processes},
(\byear{1962})
\end{bbook}
\endbibitem

\bibitem[\protect\citeauthoryear{Rubanova et~al.}{2019}]{odernn}
\begin{bchapter}
\bauthor{\bsnm{Rubanova}, \binits{Y.}},
\bauthor{\bsnm{Chen}, \binits{R.T.Q.}},
\bauthor{\bsnm{Duvenaud}, \binits{D.}}:
\bctitle{Latent odes for irregularly-sampled time series}.
In: \bbtitle{NIPS}
(\byear{2019}).
\bcomment{Arxiv: 1907.03907}
\end{bchapter}
\endbibitem

\bibitem[\protect\citeauthoryear{Yan et~al.}{2019}]{node_robustness}
\begin{bchapter}
\bauthor{\bsnm{Yan}, \binits{H.}},
\bauthor{\bsnm{Du}, \binits{J.}},
\bauthor{\bsnm{Tan}, \binits{V.Y.F.}},
\bauthor{\bsnm{Feng}, \binits{J.}}:
\bctitle{On robustness of neural ordinary differential equations}.
In: \bbtitle{ICLR}
(\byear{2019}).
\bcomment{Arxiv: 1910.05513}
\end{bchapter}
\endbibitem

\bibitem[\protect\citeauthoryear{Norcliffe et~al.}{2021}]{ndp}
\begin{bchapter}
\bauthor{\bsnm{Norcliffe}, \binits{A.}},
\bauthor{\bsnm{Bodnar}, \binits{C.}},
\bauthor{\bsnm{Day}, \binits{B.}},
\bauthor{\bsnm{Moss}, \binits{J.}},
\bauthor{\bsnm{Liò}, \binits{P.}}:
\bctitle{Neural ode processes}.
In: \bbtitle{ICLR}
(\byear{2021}).
\bcomment{Arxiv: 2103.12413}
\end{bchapter}
\endbibitem

\bibitem[\protect\citeauthoryear{Kim et~al.}{2019}]{attn_np}
\begin{bchapter}
\bauthor{\bsnm{Kim}, \binits{H.}},
\bauthor{\bsnm{Mnih}, \binits{A.}},
\bauthor{\bsnm{Schwarz}, \binits{J.}},
\bauthor{\bsnm{Garnelo}, \binits{M.}},
\bauthor{\bsnm{Ali~Eslami}, \binits{D.R.}},
\bauthor{\bsnm{Vinyals}, \binits{O.}},
\bauthor{\bsnm{Teh}, \binits{Y.W.}}:
\bctitle{Attentive neural processes}.
In: \bbtitle{ICLR}
(\byear{2019}).
\bcomment{Arxiv: 1901.05761}
\end{bchapter}
\endbibitem

\bibitem[\protect\citeauthoryear{Vaswani et~al.}{2017}]{attention_is_all_you_need}
\begin{bchapter}
\bauthor{\bsnm{Vaswani}, \binits{A.}},
\bauthor{\bsnm{Shazeer}, \binits{N.}},
\bauthor{\bsnm{Parmar}, \binits{N.}},
\bauthor{\bsnm{Uszkoreit}, \binits{J.}},
\bauthor{\bsnm{Jones}, \binits{L.}},
\bauthor{\bsnm{Gomez}, \binits{A.N.}},
\bauthor{\bsnm{Kaiser}, \binits{L.}},
\bauthor{\bsnm{Polosukhin}, \binits{I.}}:
\bctitle{Attention is all you need}.
In: \bbtitle{NIPS}
(\byear{2017}).
\bcomment{Arxiv: 1706.03762}
\end{bchapter}
\endbibitem

\bibitem[\protect\citeauthoryear{Paszke et~al.}{2019}]{pytorch}
\begin{bchapter}
\bauthor{\bsnm{Paszke}},
\bauthor{\bsnm{Adam}},
\bauthor{\bsnm{Gross}},
\bauthor{\bsnm{Sam}},
\bauthor{\bsnm{Massa}},
\bauthor{\bsnm{Francisco}},
\bauthor{\bsnm{Lerer}},
\bauthor{\bsnm{Adam}},
\bauthor{\bsnm{Bradbury}},
\bauthor{\bsnm{James}},
\bauthor{\bsnm{Chanan}},
\bauthor{\bsnm{Gregory}},
\bauthor{\bsnm{Killeen}},
\bauthor{\bsnm{Trevor}},
\bauthor{\bsnm{Lin}},
\bauthor{\bsnm{Zeming}},
\bauthor{\bsnm{Gimelshein}},
\bauthor{\bsnm{Natalia}},
\bauthor{\bsnm{Antiga}},
\bauthor{\bsnm{Luca}},
\bauthor{\bsnm{Desmaison}},
\bauthor{\bsnm{Alban}},
\bauthor{\bsnm{Kopf}},
\bauthor{\bsnm{Andreas}},
\bauthor{\bsnm{Yang}},
\bauthor{\bsnm{Edward}},
\bauthor{\bsnm{DeVito}},
\bauthor{\bsnm{Zachary}},
\bauthor{\bsnm{Raison}},
\bauthor{\bsnm{Martin}},
\bauthor{\bsnm{Tejani}},
\bauthor{\bsnm{Alykhan}},
\bauthor{\bsnm{Chilamkurthy}},
\bauthor{\bsnm{Sasank}},
\bauthor{\bsnm{Steiner}},
\bauthor{\bsnm{Benoit}},
\bauthor{\bsnm{Fang}},
\bauthor{\bsnm{Lu}},
\bauthor{\bsnm{Bai}},
\bauthor{\bsnm{Junjie}},
\bauthor{\bsnm{Chintala}},
\bauthor{\bsnm{Soumith}}:
\bctitle{Pytorch: An imperative style, high-performance deep learning library}.
In: \bbtitle{NIPS},
(\byear{2019}).
\bcomment{Arxiv: 1912.01703}
\end{bchapter}
\endbibitem

\bibitem[\protect\citeauthoryear{Nix and Weigend}{1994}]{gaussian_nloss}
\begin{bchapter}
\bauthor{\bsnm{Nix}, \binits{D.A.}},
\bauthor{\bsnm{Weigend}, \binits{A.S.}}:
\bctitle{Estimating the mean and variance of the target probability distribution}.
In: \bbtitle{ICNN}
(\byear{1994})
\end{bchapter}
\endbibitem

\bibitem[\protect\citeauthoryear{Ba et~al.}{2016}]{layer_norm}
\begin{bchapter}
\bauthor{\bsnm{Ba}, \binits{J.L.}},
\bauthor{\bsnm{Kiros}, \binits{J.R.}},
\bauthor{\bsnm{Hinton}, \binits{G.E.}}:
\bctitle{Layer normalization}.
In: \bbtitle{arXiv Preprint}
(\byear{2016}).
\bcomment{Arxiv: 1607.06450}
\end{bchapter}
\endbibitem

\bibitem[\protect\citeauthoryear{Xu et~al.}{2015}]{leaky_relu}
\begin{bchapter}
\bauthor{\bsnm{Xu}, \binits{B.}},
\bauthor{\bsnm{Wang}, \binits{N.}},
\bauthor{\bsnm{Chen}, \binits{T.}},
\bauthor{\bsnm{Li}, \binits{M.}}:
\bctitle{Empirical evaluation of rectified activations in convolution network}.
In: \bbtitle{arXiv Preprint}
(\byear{2015}).
\bcomment{Arxiv: 1505.00853}
\end{bchapter}
\endbibitem

\bibitem[\protect\citeauthoryear{Penny}{2023}]{penny2023bayesian}
\begin{botherref}
\oauthor{\bsnm{Penny}, \binits{W.}}:
Bayesian Inference, Dynamical Systems and the Brain.
\url{https://www.fil.ion.ucl.ac.uk/~wpenny/}.
Accessed: 2023-11-15
(2023)
\end{botherref}
\endbibitem

\bibitem[\protect\citeauthoryear{Liu et~al.}{2023}]{sparsetrack}
\begin{bchapter}
\bauthor{\bsnm{Liu}, \binits{Z.}},
\bauthor{\bsnm{Wang}, \binits{X.}},
\bauthor{\bsnm{Wang}, \binits{C.}},
\bauthor{\bsnm{Liu}, \binits{W.}},
\bauthor{\bsnm{Bai}, \binits{X.}}:
\bctitle{Sparsetrack: Multi-object tracking by performing scene decomposition based on pseudo-depth}.
In: \bbtitle{arXiv Preprint}
(\byear{2023}).
\bcomment{Arxiv: 2306.05238}
\end{bchapter}
\endbibitem

\bibitem[\protect\citeauthoryear{Xiao et~al.}{2023}]{motiontrack}
\begin{bchapter}
\bauthor{\bsnm{Xiao}, \binits{C.}},
\bauthor{\bsnm{Cao}, \binits{Q.}},
\bauthor{\bsnm{Zhong}, \binits{Y.}},
\bauthor{\bsnm{Lan}, \binits{L.}},
\bauthor{\bsnm{Zhang}, \binits{X.}},
\bauthor{\bsnm{Cai}, \binits{H.}},
\bauthor{\bsnm{Luo}, \binits{Z.}},
\bauthor{\bsnm{Tao}, \binits{D.}}:
\bctitle{Motiontrack: Learning motion predictor for multiple object tracking}.
In: \bbtitle{arXiv Preprint}
(\byear{2023}).
\bcomment{Arxiv: 2306.02585}
\end{bchapter}
\endbibitem

\bibitem[\protect\citeauthoryear{Zhang et~al.}{2023}]{bytev2}
\begin{bchapter}
\bauthor{\bsnm{Zhang}, \binits{Y.}},
\bauthor{\bsnm{Wang}, \binits{X.}},
\bauthor{\bsnm{Ye}, \binits{X.}},
\bauthor{\bsnm{Zhang}, \binits{W.}},
\bauthor{\bsnm{Lu}, \binits{J.}},
\bauthor{\bsnm{Tan}, \binits{X.}},
\bauthor{\bsnm{Ding}, \binits{E.}},
\bauthor{\bsnm{Sun}, \binits{P.}},
\bauthor{\bsnm{Wang}, \binits{J.}}:
\bctitle{Bytetrackv2: 2d and 3d multi-object tracking by associating every detection box}.
In: \bbtitle{Combined Ophthalmic Research Rotterdam}
(\byear{2023}).
\bcomment{Arxiv: 2303.15334}
\end{bchapter}
\endbibitem

\bibitem[\protect\citeauthoryear{Cui et~al.}{2023}]{sportsmot}
\begin{bchapter}
\bauthor{\bsnm{Cui}, \binits{Y.}},
\bauthor{\bsnm{Zeng}, \binits{C.}},
\bauthor{\bsnm{Zhao}, \binits{X.}},
\bauthor{\bsnm{Yang}, \binits{Y.}},
\bauthor{\bsnm{Wu}, \binits{G.}},
\bauthor{\bsnm{Wang}, \binits{L.}}:
\bctitle{Sportsmot: A large multi-object tracking dataset in multiple sports scenes}.
In: \bbtitle{CVPR}
(\byear{2023}).
\bcomment{Arxiv: 2304.05170}
\end{bchapter}
\endbibitem

\bibitem[\protect\citeauthoryear{Fan et~al.}{2018}]{lasot1}
\begin{bchapter}
\bauthor{\bsnm{Fan}, \binits{H.}},
\bauthor{\bsnm{Lin}, \binits{L.}},
\bauthor{\bsnm{Yang}, \binits{F.}},
\bauthor{\bsnm{Chu}, \binits{P.}},
\bauthor{\bsnm{Deng}, \binits{G.}},
\bauthor{\bsnm{Yu}, \binits{S.}},
\bauthor{\bsnm{Bai}, \binits{H.}},
\bauthor{\bsnm{Xu}, \binits{Y.}},
\bauthor{\bsnm{Liao}, \binits{C.}},
\bauthor{\bsnm{Ling}, \binits{H.}}:
\bctitle{Lasot: A high-quality benchmark for large-scale single object tracking}.
In: \bbtitle{CVPR}
(\byear{2018}).
\bcomment{Arxiv: 1809.07845}
\end{bchapter}
\endbibitem

\bibitem[\protect\citeauthoryear{Fan et~al.}{2020}]{lasot2}
\begin{bchapter}
\bauthor{\bsnm{Fan}, \binits{H.}},
\bauthor{\bsnm{Bai}, \binits{H.}},
\bauthor{\bsnm{Lin}, \binits{L.}},
\bauthor{\bsnm{Yang}, \binits{F.}},
\bauthor{\bsnm{Chu}, \binits{P.}},
\bauthor{\bsnm{Deng}, \binits{G.}},
\bauthor{\bsnm{Yu}, \binits{S.}},
\bauthor{\bsnm{Harshit}, \binits{M.H.}},
\bauthor{\bsnm{Liu}, \binits{J.}},
\bauthor{\bsnm{Xu}, \binits{Y.}},
\bauthor{\bsnm{Liao}, \binits{C.}},
\bauthor{\bsnm{Yuan}, \binits{L.}},
\bauthor{\bsnm{Ling}, \binits{H.}}:
\bctitle{Lasot: A high-quality large-scale single object tracking benchmark}.
In: \bbtitle{IJCV}
(\byear{2020}).
\bcomment{Arxiv: 2009.03465}
\end{bchapter}
\endbibitem

\bibitem[\protect\citeauthoryear{Maggiolino et~al.}{2023}]{deepocsort}
\begin{bchapter}
\bauthor{\bsnm{Maggiolino}, \binits{G.}},
\bauthor{\bsnm{Ahmad}, \binits{A.}},
\bauthor{\bsnm{Jinkun~Cao}, \binits{K.K.}}:
\bctitle{Deep oc-sort: Multi-pedestrian tracking by adaptive re-identification}.
In: \bbtitle{ICIP}
(\byear{2023}).
\bcomment{Arxiv: 2302.11813}
\end{bchapter}
\endbibitem

\bibitem[\protect\citeauthoryear{Qin et~al.}{2023}]{motiontrack_byte_cmc}
\begin{bchapter}
\bauthor{\bsnm{Qin}, \binits{Z.}},
\bauthor{\bsnm{Zhou}, \binits{S.}},
\bauthor{\bsnm{Wang}, \binits{L.}},
\bauthor{\bsnm{Duan}, \binits{J.}},
\bauthor{\bsnm{Hua}, \binits{G.}},
\bauthor{\bsnm{Tang}, \binits{W.}}:
\bctitle{Motiontrack: Learning robust short-term and long-term motions for multi-object tracking}.
In: \bbtitle{CVPR}
(\byear{2023}).
\bcomment{Arxiv: 2303.10404}
\end{bchapter}
\endbibitem

\bibitem[\protect\citeauthoryear{Runge}{1895}]{runge_kutta}
\begin{bbook}
\bauthor{\bsnm{Runge}, \binits{C.}}:
\bbtitle{Über die Numerische Auflösung Von Differentialgleichungen},
(\byear{1895})
\end{bbook}
\endbibitem

\bibitem[\protect\citeauthoryear{Cao et~al.}{2020}]{hota}
\begin{bchapter}
\bauthor{\bsnm{Cao}, \binits{J.}},
\bauthor{\bsnm{Pang}, \binits{J.}},
\bauthor{\bsnm{Xinshuo~Weng}, \binits{R.K.}},
\bauthor{\bsnm{Kitani}, \binits{K.}}:
\bctitle{Hota: A higher order metric for evaluating multi-object tracking}.
In: \bbtitle{IJCV}
(\byear{2020}).
\bcomment{Arxiv: 2009.07736}
\end{bchapter}
\endbibitem

\bibitem[\protect\citeauthoryear{Singh et~al.}{2019}]{snp}
\begin{bchapter}
\bauthor{\bsnm{Singh}, \binits{G.}},
\bauthor{\bsnm{Yoon}, \binits{J.}},
\bauthor{\bsnm{Son}, \binits{Y.}},
\bauthor{\bsnm{Ahn}, \binits{S.}}:
\bctitle{Sequential neural processes}.
In: \bbtitle{NIPS}
(\byear{2019}).
\bcomment{Arxiv: 1906.10264}
\end{bchapter}
\endbibitem

\bibitem[\protect\citeauthoryear{Revach et~al.}{2019}]{kalman_net}
\begin{bchapter}
\bauthor{\bsnm{Revach}, \binits{G.}},
\bauthor{\bsnm{Shlezinger}, \binits{N.}},
\bauthor{\bsnm{Ni}, \binits{X.}},
\bauthor{\bsnm{Escoriza}, \binits{A.L.}},
\bauthor{\bsnm{Sloun}, \binits{R.J.G.}},
\bauthor{\bsnm{Eldar}, \binits{Y.C.}}:
\bctitle{Kalmannet: Neural network aided kalman filtering for partially known dynamics}.
In: \bbtitle{IEEE TSP}
(\byear{2019}).
\bcomment{Arxiv: 2107.10043}
\end{bchapter}
\endbibitem

\bibitem[\protect\citeauthoryear{Brouwer et~al.}{2019}]{gru_ode_bayes}
\begin{bchapter}
\bauthor{\bsnm{Brouwer}, \binits{E.D.}},
\bauthor{\bsnm{Simm}, \binits{J.}},
\bauthor{\bsnm{Adam~Arany}, \binits{Y.M.}}:
\bctitle{Gru-ode-bayes: Continuous modeling of sporadically-observed time series}.
In: \bbtitle{NIPS}
(\byear{2019}).
\bcomment{Arxiv: 1905.12374}
\end{bchapter}
\endbibitem

\bibitem[\protect\citeauthoryear{Goodfellow et~al.}{2016}]{Goodfellow-et-al-2016}
\begin{bbook}
\bauthor{\bsnm{Goodfellow}, \binits{I.}},
\bauthor{\bsnm{Bengio}, \binits{Y.}},
\bauthor{\bsnm{Courville}, \binits{A.}}:
\bbtitle{Deep Learning},
(\byear{2016}).
\bcomment{\url{http://www.deeplearningbook.org}}
\end{bbook}
\endbibitem

\bibitem[\protect\citeauthoryear{P et~al.}{2014}]{adam}
\begin{bchapter}
\bauthor{\bsnm{P}, \binits{D.}},
\bauthor{\bsnm{Kingma}},
\bauthor{\bsnm{Ba}, \binits{J.}}:
\bctitle{Adam: A method for stochastic optimization}.
In: \bbtitle{ICLR}
(\byear{2014})
\end{bchapter}
\endbibitem

\bibitem[\protect\citeauthoryear{Loshchilov and Hutter}{2019}]{adamw}
\begin{bchapter}
\bauthor{\bsnm{Loshchilov}, \binits{I.}},
\bauthor{\bsnm{Hutter}, \binits{F.}}:
\bctitle{Decoupled weight decay regularization}.
In: \bbtitle{ICLR}
(\byear{2019})
\end{bchapter}
\endbibitem

\bibitem[\protect\citeauthoryear{Ghosh et~al.}{1991}]{weight_decay}
\begin{bchapter}
\bauthor{\bsnm{Ghosh}, \binits{A.}},
\bauthor{\bsnm{Behl}, \binits{H.S.}},
\bauthor{\bsnm{Dupont}, \binits{E.}},
\bauthor{\bsnm{Torr}, \binits{P.H.S.}},
\bauthor{\bsnm{Namboodiri}, \binits{V.}}:
\bctitle{A simple weight decay can improve generalization}.
In: \bbtitle{NIPS}
(\byear{1991})
\end{bchapter}
\endbibitem

\bibitem[\protect\citeauthoryear{Huang et~al.}{2023}]{deep_eiou}
\begin{bchapter}
\bauthor{\bsnm{Huang}, \binits{H.-W.}},
\bauthor{\bsnm{Yang}, \binits{C.-Y.}},
\bauthor{\bsnm{Sun}, \binits{J.}},
\bauthor{\bsnm{Kim}, \binits{P.-K.}},
\bauthor{\bsnm{Kim}, \binits{K.-J.}},
\bauthor{\bsnm{Lee}, \binits{K.}},
\bauthor{\bsnm{Huang}, \binits{C.-I.}},
\bauthor{\bsnm{Hwang}, \binits{J.-N.}}:
\bctitle{Iterative scale-up expansioniou and deep features association for multi-object tracking in sports}.
In: \bbtitle{WACV RWS Workshop}
(\byear{2023}).
\bcomment{arxiv:2306.13074}
\end{bchapter}
\endbibitem

\bibitem[\protect\citeauthoryear{Lin et~al.}{2014}]{coco}
\begin{bchapter}
\bauthor{\bsnm{Lin}, \binits{T.-Y.}},
\bauthor{\bsnm{Maire}, \binits{M.}},
\bauthor{\bsnm{Belongie}, \binits{S.}},
\bauthor{\bsnm{Bourdev}, \binits{L.}},
\bauthor{\bsnm{Girshick}, \binits{R.}},
\bauthor{\bsnm{Hays}, \binits{J.}},
\bauthor{\bsnm{Perona}, \binits{P.}},
\bauthor{\bsnm{Ramanan}, \binits{D.}},
\bauthor{\bsnm{Zitnick}, \binits{C.L.}},
\bauthor{\bsnm{Dollár}, \binits{P.}}:
\bctitle{Microsoft coco: Common objects in context}.
In: \bbtitle{ECCV}
(\byear{2014})
\end{bchapter}
\endbibitem

\end{thebibliography}

\end{document}